\documentclass{article}

\PassOptionsToPackage{numbers}{natbib}
\usepackage[main, final]{neurips_2026}

\usepackage[utf8]{inputenc} %
\usepackage[T1]{fontenc}    %
\usepackage{hyperref}       %
\usepackage{url}            %
\usepackage{booktabs}       %
\usepackage{amsfonts}       %
\usepackage{nicefrac}       %
\usepackage{microtype}      %
\usepackage{xcolor}         %
\usepackage{amsmath}
\usepackage{graphicx}
\newcommand{\algname}{ManiFusion}
\newcommand{\algacronym}{MF}
\usepackage{multirow}
\usepackage{adjustbox}
\usepackage{placeins} %

\title{ManiFusion: Unlocking the High-Throughput Generation via Superposition in Manifold Space}

\author{%
  \textbf{Minkyu Kim}$^{1}$ \quad \textbf{Baekseung Kim}$^{1}$ \quad \textbf{Junhoo Lee}$^{1}$ \quad
  \textbf{Jangho Kim}$^{2,}$\thanks{Co-corresponding authors: Jangho Kim and Nojun Kwak.} \quad \textbf{Nojun Kwak}$^{1,*}$ \\
  $^{1}$Seoul National University \quad $^{2}$Kookmin University \\
  {\small\texttt{\{minkyu.kim,baekseung.kim,mrjunoo,nojunk\}@snu.ac.kr}} \\
  {\small\texttt{jangho.kim@kookmin.ac.kr}}
}

\begin{document}

\maketitle
\setcounter{footnote}{0}

\begin{abstract}
Diffusion models achieve high-quality image generation but remain computationally expensive due to iterative denoising across many timesteps. Existing acceleration methods mainly reduce per-sample latency, but gains along this axis are becoming saturated. We instead explore an orthogonal throughput-oriented direction by processing multiple samples within a single denoising evaluation. To this end, we propose ManiFusion, a backbone-agnostic framework for multi-sample generation through fused-space computation. ManiFusion jointly processes multiple samples by fusing them before the denoising backbone and separating them afterward through unfusion. We provide a geometric perspective on how multiple samples can share a fused representation while retaining the information needed to guide their individual generation trajectories. To better align fusion allocation with denoising stages, ManiFusion uses Trajectory-Fusion Scheduling (TFS), which dynamically adjusts fusion factors throughout sampling. ManiFusion generalizes across diffusion families, including both U-Net and transformer backbones in pixel and latent spaces. We further show that ManiFusion composes with existing acceleration methods and reaches throughput regimes unattainable by prior approaches with minimal quality degradation.
\end{abstract}

\section{Introduction}

Diffusion models have become the dominant framework for high-fidelity image generation by progressively refining noise into structured samples.~\cite{ho2020denoising, song2020denoising, rombach2022high} However, this strong generation quality comes at the cost of iterative denoising across many timesteps, making sampling inherently sequential and limiting throughput.\footnote{\label{fn:throughput}$\mathrm{Throughput} = \frac{\text{images generated}}{\text{total inference time}} = \frac{\text{images per forward pass}}{\text{inference time per forward pass}}$.} Prior work has made substantial progress by reducing the per-sample cost of diffusion sampling. One line of work reduces the number of denoising evaluations.~\cite{song2020denoising, song2023consistency, salimans2022progressive, lu2022dpm, lu2025dpm, zhang2022fast, zhou2024fast, zhou2024simple} Another line of work reduces the cost of individual denoising evaluations through feature reuse or caching.~\cite{ma2024deepcache, li2024faster} Together, these approaches have substantially improved the efficiency--quality trade-off of diffusion sampling in practice.

Despite these advances, further acceleration along the same per-sample axis is increasingly difficult. As denoising evaluations decrease, each remaining step must approximate more of the trajectory, making the dynamics more sensitive to approximation error.~\cite{xia2023towards, yang2023denoising} Caching methods reuse intermediate features across adjacent timesteps, but wider cache intervals amplify mismatch from fully recomputed trajectories.~\cite{ma2024deepcache} When the two approaches are combined, reduced-step sampling makes each update more consequential while caching supplies features from increasingly mismatched trajectory states, limiting further gains without additional quality loss overall.

\begin{figure}[t]
    \centering
    \includegraphics[width=0.8\linewidth]{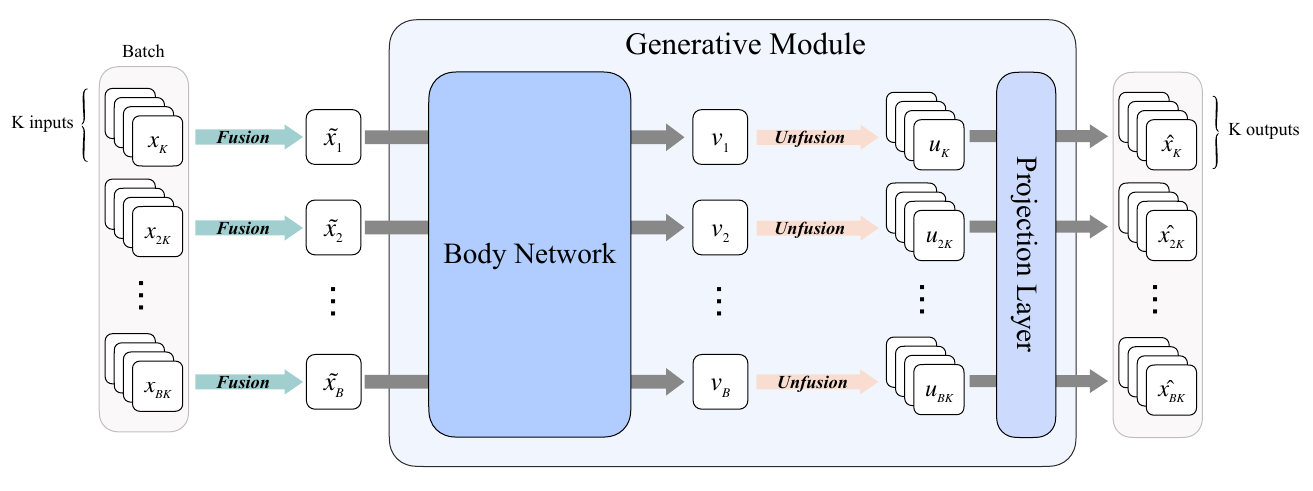}
    \caption{Illustration of the \algname{} framework. $K \times B$ inputs is fused into $B$ representations, jointly processed by a shared denoising backbone, then unfused via the \textit{unfusion} module into $K$ representations per group, and finally passed through a projection layer to produce $K$ outputs.
  }
    \label{fig:fusebatch}
\end{figure}

As gains along the per-sample axis approach saturation, we instead increase throughput by processing more samples within each denoising evaluation. To this end, we propose \textit{\algname{}}, a lightweight framework for multi-sample generation through fused-space computation. \algname{} processes multiple samples within a shared denoising backbone by fusing their representations before the backbone and separating them afterward through index-conditioned unfusion. Specifically, \algname{} realizes multi-sample computation through \textit{Trajectory-Fusion Scheduling} (TFS), which uses larger fusion factors in stable regions and smaller factors in sensitive stages.

However, generating multiple samples through a single fused representation presents a nontrivial challenge. Each sample needs a distinct denoising prediction, yet fusion combines their states into a representation the size of a single input. Naive superposition can entangle their information~\cite{zhang2017mixup,verma2019manifold} and compromise the denoising dynamics governing their individual generation trajectories.~\cite{song2020denoising,lipman2022flow} We examine how shared computation can preserve these distinct predictions through a geometric perspective under the manifold hypothesis.~\cite{olshausen1996emergence,simoncelli2001natural,carlsson2009topology,chapelle2006semi} Low-dimensional data structure provides room for samples to coexist, while denoising requires their prediction targets to remain recoverable. Our analysis relates fusion-induced information loss to prediction error, supported by empirical studies of sample separation and denoising behavior.

To demonstrate that ManiFusion generalizes across diverse generative settings, we validate ManiFusion across major diffusion model families, including pixel-space U-Nets, diffusion transformers, and latent diffusion models.~\cite{ronneberger2015u, peebles2023scalable, rombach2022high} Our experiments cover both unconditional and conditional generation across broad resolution ranges from CIFAR-10 to ImageNet. We further show that ManiFusion composes naturally with existing acceleration methods, including step-reduction samplers and caching-based acceleration.~\cite{lu2022dpm, song2023consistency, ma2024deepcache} By operating along a different throughput axis, ManiFusion reaches higher-throughput regimes beyond trajectory shortening or per-step acceleration alone.

\section{Method}
\label{sec:Method}

We propose \algname{}, a backbone-agnostic framework for multi-sample generation through fused-space computation. Since multi-sample generation requires jointly processing multiple trajectories while preserving sample-wise prediction targets, we first establish \algname{} through dual-space inference (Sec.~\ref{subsec:Formulation}). In this framework, multiple denoising evaluations are replaced by a single fused computation followed by index-conditioned recovery. We then extend \algname{} across diverse diffusion backbones and sampling paradigms through Trajectory-Fusion Scheduling (TFS), which dynamically allocates the fusion factor according to trajectory dynamics (Sec.~\ref{subsec:practical_instantiations}). Finally, we develop a geometric intuition for why shared computation can support distinct generation trajectories and what information fusion must preserve to make this possible (Sec.~\ref{subsec:geometric_analysis}).

\subsection{\algname{}}
\label{subsec:Formulation}

\algname{} enables multiple samples to share computation through fusion into a common representation and unfusion into individual outputs. We first formulate this principle for denoiser-level fusion. Diffusion models operate on an ambient representation space $\mathbb{R}^{D}$, corresponding to either pixel or learned latent space depending on the model, with denoising (or flow) operator $f_t:\mathbb{R}^{D}\rightarrow\mathbb{R}^{D}$ at timestep $t$. Given $K$ inputs $(x_t^1,\dots,x_t^K)\in(\mathbb{R}^{D})^K$, independent denoising aims to predict the sample-wise targets as
$
\bigl(f_t(x_t^1),\dots,f_t(x_t^K)\bigr)
\approx
(y_t^1,\dots,y_t^K).
$ Both independent and fused models are trained against these targets $y_t^i$, with the fused model's predictions denoted by $\hat y_t^i$.

The key idea of \algname{} is to replace these independent evaluations with a single computation in a shared fused space. To enable this, we first define a fusion operator $F_K:(\mathbb{R}^{D})^K\rightarrow\mathbb{R}^{D}$ that combines representations through index-conditioned summation across sample slots
\begin{equation}
z_t
=
F_K(x_t^1,\dots,x_t^K)
=
\sum_{i=1}^{K}\psi_{K,i}(x_t^i),
\end{equation}
where $\psi_{K,i}:\mathbb{R}^{D}\rightarrow\mathbb{R}^{D}$ denotes the encoding function associated with the $i$-th sample slot.

Once the inputs are fused into a shared representation, computation proceeds through a fused-space operator
$\tilde f_{t,K}:\mathbb{R}^{D}\rightarrow\mathbb{R}^{D}$,
which is designed to approximate the effect of applying the independent operator $f_t$ to each trajectory within the fused space. From this perspective, \algname{} can be viewed as a dual-space computation framework, where generative dynamics are carried out through the coupled original sample space and fused shared space. After fused processing, individual predictions are recovered by the following index-conditioned unfusion operator
\begin{equation}
U_K(z)_i = \varphi_{K,i}(z), \quad i=1,\dots,K,
\end{equation}
where $\varphi_{K,i}:\mathbb{R}^{D}\rightarrow\mathbb{R}^{D}$ denotes the decoding function for the $i$-th sample slot, including the final prediction head in this abstract notation. Thus, \algname{} aims to satisfy
\begin{equation}
\label{eq:dual_approx}
U_K\!\left(
\tilde f_{t,K}
\left(
F_K(x_t^1,\dots,x_t^K)
\right)
\right)
\approx
\bigl(f_t(x_t^1),\dots,f_t(x_t^K)\bigr).
\end{equation}

The following diagram illustrates how independent and fused computation aim to predict the same sample-wise targets.
\begin{equation}
\begin{array}{ccc}
z_t
&
\xrightarrow{\hspace{0.7cm} \tilde f_{t,K} \hspace{0.7cm}}
&
\tilde y_t
\\[0.1cm]
\big\uparrow{\scriptstyle F_K}
&
&
\big\downarrow{\scriptstyle U_K}
\\[0.1cm]
(x_t^1,\dots,x_t^K)
&
\xrightarrow{\hspace{0.7cm} (f_t,\dots,f_t) \hspace{0.7cm}}
&
(y_t^1,\dots,y_t^K)
\end{array}
\end{equation}
where $z_t = F_K(x_t^1,\dots,x_t^K)$ and $\tilde y_t = \tilde f_{t,K}(z_t)$.

\paragraph{\textbf{Trajectory-Fusion Scheduling.}}

For these denoiser-level variants, \algname{} uses Trajectory-Fusion Scheduling (TFS), which allocates the fusion factor according to the local behavior of the generative trajectory. This is motivated by the observation that the approximation quality in Eq.~\ref{eq:dual_approx} varies throughout the diffusion process, since the stability of dual-space computation depends on the local trajectory behavior. We therefore define the fusion factor $K_t=\kappa(t)$ as a function of timestep $t$. At each timestep, samples are grouped into tuples of size $K_t$, and fused computation is applied to each group as
$z_t = F_{K_t}(x_t^1,\dots,x_t^{K_t})$ and $(\hat y_t^1,\dots,\hat y_t^{K_t}) = U_{K_t}(\tilde f_{t,K_t}(z_t))$. By allocating fusion factors according to trajectory dynamics, \algname{} balances computational efficiency and sample fidelity across denoising stages without changing the backbone.

\subsection{Instantiation Across Architectures and Generative Flows}
\label{subsec:practical_instantiations}

To examine whether fused computation generalizes across model designs, we instantiate \algname{} in U-Nets, diffusion transformers, and autoencoders, covering convolutional and attention-based backbones as well as denoiser-level and latent-representation sharing. We further consider DDPM, DDIM, and Flow Matching to study how this principle applies across different generative dynamics.

\paragraph{\textbf{Backbone restructuring}}

To maximize shared computation across multiple trajectories, \algname{} restructures the denoiser such that most computation is performed in the fused space. Let $\mathcal{B}$ denote the heavy backbone body of the denoiser and $\mathcal{H}$ its final projection head. Instead of repeatedly evaluating $\mathcal{B}$ for each trajectory independently, \algname{} places the fusion module $F$ before $\mathcal{B}$ and the unfusion module $U$ after $\mathcal{B}$, allowing the dominant backbone computation to be shared across fused trajectories. This yields the following computational structure.
\begin{equation}
{x_t^{(i,j)}}_{j=1}^{K}
;\xrightarrow{F};
\tilde{x}_t^{(i)}
;\xrightarrow{\mathcal{B}};
\tilde{v}_t^{(i)}
;\xrightarrow{U(\cdot,j)};
u_t^{(i,j)}
;\xrightarrow{\mathcal{H}};
\hat{\epsilon}_t^{(i,j)}.
\label{eq:fusebatch_template}
\end{equation}
Here $i$ indexes a fusion group and $j\in\{1,\ldots,K\}$ indexes a sample slot within that group.

\paragraph{\textbf{\algacronym{}-UNet}}

For pixel-space models such as DDPM, we split the U-Net at its final convolutional layer, treating the multi-scale convolutional backbone as the shared body and the output convolution as the head. Fusion uses lightweight $3\times3$ convolutional blocks per index, while unfusion uses pointwise convolutions. Because the body output already encodes rich spatial features~\cite{ronneberger2015u}, simple channel-wise separation suffices without modifying the U-Net internals.

\paragraph{\textbf{\algacronym{}-DiT}}

For transformer-based models such as DiT, inputs have shape $B \times K \times L \times d$, where $L$ is the number of patch tokens. To treat the whole transformer stack as the shared body, we split the denoiser at the final linear projection. Fusion and unfusion are implemented as index-conditioned fully connected layers applied identically across patch positions $\ell\in\{1,\dots,L\}$, sharing parameters to preserve token-wise structure and permutation sensitivity across all patch tokens.

\paragraph{\textbf{\algacronym{}-AE}}

For latent diffusion models, fusion is applied at the representation level of the autoencoder rather than within the denoising trajectory. We extend \algname{} by inserting fusion before encoding and unfusion after decoding, effectively operating on a shared latent representation. During autoencoder training, multiple inputs are fused prior to encoding and reconstructed through a shared latent. At generation time, a shared latent is denoised and decoded jointly. Unfusion is then applied at the output to recover individual samples. Because fusion operates at the representation level, $K$ remains fixed during sampling.

\paragraph{\textbf{Trajectory-dependent fusion allocation}}

To show that trajectory-dependent fusion naturally generalizes across both score-based diffusion and velocity-based generative flows, we apply TFS by switching between independently trained fixed-$K$ models during inference. For DDPM, larger fusion factors are assigned to early noisy stages and smaller factors to later stages, consistent with the coarse-to-fine behavior of diffusion trajectories~\cite{karras2022elucidating,yang2023denoising}. Specifically, we allocate 50\% of timesteps to $K{=}4$, 25\% to $K{=}2$, and 25\% to $K{=}1$.\footnote{This yields an idealized $2\times$ throughput since the effective cost is $\tfrac{1}{4} \times 0.5 + \tfrac{1}{2} \times 0.25 + 1 \times 0.25 = 0.5$.} DDIM~\cite{song2020denoising} uses the same allocation. For Flow Matching~\cite{lipman2022flow}, we instead assign larger fusion factors to regions with smaller scheduler intervals, where the same velocity prediction error contributes a smaller displacement error to each update (Sec.~\ref{subsec:geometric_analysis}). Finally, we combine \algname{} with step-reduction samplers and per-step acceleration methods such as DeepCache~\cite{ma2024deepcache}, achieving additional throughput gains with only slight quality degradation.

\subsection{Geometric Analysis}
\label{subsec:geometric_analysis}
Fusing multiple samples reduces computation, but how can a single shared representation support the distinct predictions needed to generate each sample? To answer this question, we examine which differences between inputs must remain distinguishable after fusion for their predictions to be recovered. The low-dimensional structure of clean data provides a geometric basis for preserving these distinctions within a shared space. We then extend this reasoning to noisy inputs, where the information needed for prediction changes along the denoising trajectory, motivating adaptive allocation of shared computation.

\paragraph{Prediction information retained by fusion.}
For shared computation to support individual generation trajectories, fusion must preserve the information needed for the original sample-wise predictions. The training targets and squared-error objective remain unchanged, so the key question is whether these predictions can still be obtained through the shared representation. At timestep $t$, let $X_t=(x_t^1,\ldots,x_t^K)$ and $Y_t=(y_t^1,\ldots,y_t^K)$ denote the input and target tuples in $\mathbb R^{KD}$. Depending on the formulation, each target is noise~\cite{ho2020denoising}, velocity~\cite{lipman2022flow}, or a clean state~\cite{salimans2022progressive}. Fusion maps the inputs to $Z_t=F_K(X_t)\in\mathbb R^D$, from which $G_t$ produces the sample-wise predictions $\widehat Y_t=G_t(Z_t)$. Here $G_t$ comprises the shared backbone, index-conditioned unfusion, and output head.

To make this requirement precise, we use the optimal prediction from the full input tuple as a reference. Under squared loss, this is the conditional mean (Appendix~\ref{sec:target_details}, Eq.~\ref{eq:conditional_projection})
\[
m_t(X_t):=\mathbb E_t[Y_t\mid X_t],
\]
where $\mathbb E_t$ denotes expectation at fixed $t$ and available conditioning, such as class labels. If the shared representation retains the information needed to recover this prediction, there exists a predictor $g_t$ such that $m_t(X_t)=g_t(Z_t)$ almost surely. Fusion can then attain the same minimum prediction error as the full-input reference. This condition identifies what fusion must preserve to share computation without introducing additional prediction error.

\paragraph{Prediction ambiguity under superposition.}
This criterion fails when tuples sharing a representation require different predictions. For a representation $z$, define its \emph{fiber} as $F_K^{-1}(z)=\{X:F_K(X)=z\}$. If the optimal targets vary across this set, the fused predictor must compromise between them. The same squared-loss argument gives $g_t^*(Z_t)=\mathbb E_t[Y_t\mid Z_t]=\mathbb E_t[m_t(X_t)\mid Z_t]$ by iterated expectation, since $Z_t=F_K(X_t)$. The star denotes the fused optimum.

Define the minimum full-input risk as $R_{\mathrm{full},t}=\mathbb E_t\|Y_t-m_t(X_t)\|^2$ and the minimum fused-input risk as $R_{F,t}=\mathbb E_t\|Y_t-g_t^*(Z_t)\|^2$. Their difference defines the \emph{fusion ambiguity} as
\begin{equation}
\Delta_t(K):=R_{F,t}-R_{\mathrm{full},t}
=\mathbb E_t\|m_t(X_t)-g_t^*(Z_t)\|^2.
\label{eq:main_ambiguity}
\end{equation}
Thus collisions cause an irreducible prediction error when they hide differences in the required predictions. Agreement within each fiber makes this term vanish.

The learned predictor may also differ from $g_t^*$. Let $\theta$ denote its parameters and $R_{\theta,F,t}=\mathbb E_t\|Y_t-G_t(Z_t)\|^2$ its risk. Separating this prediction error from fusion ambiguity gives
\begin{equation}
R_{\theta,F,t}-R_{\mathrm{full},t}
=\underbrace{\Delta_t(K)}_{\text{information lost by fusion}}
+\underbrace{\mathbb E_t\|g_t^*(Z_t)-G_t(Z_t)\|^2}_{\text{prediction error on retained information}}.
\label{eq:main_risk}
\end{equation}
This decomposition separates the additional error into information lost through fusion and prediction error on the retained information. Appendix~\ref{sec:target_details}, Eq.~\ref{eq:residual_decomposition}, derives both identities. Joint training can therefore reduce harmful collisions through $F_K$ and improve recovery through $G_t$.

\paragraph{Manifold capacity and recoverable sample variations.}
To understand how several samples can fit in one representation, consider clean-state prediction, $Y_t=X_0$. At the noiseless endpoint $t=0$, $m_0(X_0)=X_0$, so preserving the prediction means recovering the clean tuple $X_0$. Under the manifold hypothesis, each clean sample lies on a smooth $d$-dimensional manifold $\mathcal M\subset\mathbb R^D$ and needs only $d$ independent local coordinates. A tuple of $K$ independently varying samples lies on the product manifold $\mathcal M^K$, with $Kd$ such coordinates.

Recovering these coordinates requires an ideal smooth decoder satisfying $g_0(F_K(X_0))=X_0$. Let $T_{X_0}\mathcal M^K$ be the tangent space of allowed local sample variations, and $J_{\mathcal M}h$ the derivative of a map $h$ restricted to these variations. Its rank counts the number of independent directions preserved by that map. Differentiating the recovery identity (Appendix~\ref{sec:target_details}, Eq.~\ref{eq:tangent_recovery}) gives
\begin{equation}
Kd=\operatorname{rank}(J_{\mathcal M}m_0)
\le\operatorname{rank}(J_{\mathcal M}F_K)\le D.
\label{eq:main_capacity}
\end{equation}
Thus smooth clean-tuple recovery requires $D\ge Kd$. This explains the room for sharing when $d\ll D$, since several samples' intrinsic coordinates can fit within one ambient representation. For \algacronym{}-AE, the same argument uses $D_{\mathrm{AE}}$, the dimension of its shared latent representation.

The capacity condition provides enough dimensions to represent a clean tuple, but preserving its information also requires separating the contributions of different slots. If changes in two slots map to the same direction, they can cancel under summation, preventing recovery even when $D\ge Kd$. This motivates the index-conditioned maps $\psi_{K,i}$, which allow each slot to encode its variations in different directions. Training against individual prediction targets encourages the fused representation to retain these distinctions. When the resulting directions are linearly independent on the clean manifold, the fusion map has full tangent rank and admits a local inverse (Appendix~\ref{sec:target_details}). For recovery to be stable, these distinctions must also remain sufficiently large. The local bound $\|F_K(X)-F_K(X')\|\ge c\|X-X'\|$ with $c>0$ quantifies this separation for nearby clean tuples, limiting inverse amplification to $1/c$. A small $c$ therefore indicates sensitive recovery even when a local inverse exists. Together, capacity and local separation explain how a shared representation can support recoverable sample-wise information, motivating the toy, response, and training analyses below.

\begin{table*}[t]
    \centering
    \caption{Unified comparison of Vanilla, \algname{} w.o.\ TFS, and \algname{} across three sampling frameworks (DDPM, DDIM (T=50), FM) and DiT on CIFAR-10 and CelebA. \algname{} w.o.\ TFS uses a fixed fusion factor with \(K=2\) throughout the denoising trajectory, while \algname{} dynamically allocates the fusion factor via TFS. DiT is evaluated with class-conditional generation on CIFAR-10 only, while all other models are unconditional. Best results are in \textbf{bold}, and second-best results are \underline{underlined}.}
    \vspace{1mm}
    \label{tab:core}

    \setlength{\tabcolsep}{3.5pt}
    \renewcommand{\arraystretch}{1.08}

    \begin{adjustbox}{max width=0.85\linewidth}
    \begin{tabular}{l|l|cc|c|ccccc}
        \toprule
        \multicolumn{10}{c}{\textbf{CIFAR-10}} \\
        \midrule
        Method & Variant & Throughput$\uparrow$ & MAC$\downarrow$ & Parameter & FID$\downarrow$ & sFID$\downarrow$ & IS$\uparrow$ & Precision$\uparrow$ & Recall$\uparrow$ \\
        \midrule

        \multirow{3}{*}{DDPM}
          & Vanilla & 1.71 & 6.07T & 35.7M & \textbf{3.41} & \underline{4.40} & \underline{9.20} & \textbf{0.68} & \underline{0.58} \\
          & Ours (-TFS) & \underline{3.17} ($\times1.85$) & \underline{3.07T} (50.6\%) & +63.1K (0.18\%) & 5.93 & 4.58 & 8.67 & 0.63 & \underline{0.58} \\
          & \textbf{Ours} & \textbf{3.31} ($\times1.94$) & \textbf{3.06T} (50.4\%) & +102.0K (0.29\%) & \underline{3.88} & \textbf{4.27} & \textbf{9.29} & \underline{0.66} & \textbf{0.60} \\
        \midrule

        \multirow{3}{*}{DDIM (T=50)}
          & Vanilla & 35.47 & 303.44G & 35.7M & \textbf{7.99} & \underline{6.66} & \textbf{8.71} & \textbf{0.64} & 0.55 \\
          & Ours (-TFS) & \underline{65.39} ($\times1.84$) & \underline{153.44G} (50.6\%) & +63.1K (0.18\%) & 13.34 & 7.16 & 8.11 & 0.59 & \underline{0.56} \\
          & \textbf{Ours} & \textbf{66.80} ($\times1.88$) & \textbf{152.87G} (50.4\%) & +102.0K (0.29\%) & \underline{10.35} & \textbf{6.29} & \underline{8.58} & \underline{0.60} & \textbf{0.57} \\
        \midrule

        \multirow{3}{*}{FM}
          & Vanilla & 9.60 & 124.55G & 55.7M & \textbf{2.74} & \textbf{3.84} & \textbf{9.47} & \textbf{0.65} & \textbf{0.63} \\
          & Ours (-TFS) & \underline{18.22} ($\times1.90$) & \underline{62.89G} (50.5\%) & +185.9K (0.33\%) & 6.72 & 4.50 & \underline{8.76} & \underline{0.62} & \underline{0.60} \\
          & \textbf{Ours} & \textbf{18.29} ($\times1.91$) & \textbf{62.70G} (50.3\%) & +323.4K (0.58\%) & \underline{4.10} & \underline{3.95} & 8.72 & 0.61 & 0.59 \\
        \midrule

        \multirow{3}{*}{DiT}
          & Vanilla & 8.25 & 217.00G & 32.48M & \textbf{10.58} & \underline{7.37} & \textbf{10.17} & \textbf{0.75} & \textbf{0.40} \\
          & Ours (-TFS) & \underline{15.86} ($\times1.92$) & \underline{114.25G} (52.6\%) & +1.04M (3.02\%) & 11.13 & \textbf{6.45} & 9.50 & 0.72 & \textbf{0.40} \\
          & \textbf{Ours} & \textbf{15.95} ($\times1.93$) & \textbf{113.84G} (52.5\%) & +1.63M (5.02\%) & \underline{10.97} & 7.71 & \underline{9.60} & \underline{0.73} & \underline{0.37} \\

        \midrule
        \multicolumn{10}{c}{\textbf{CelebA}} \\
        \midrule
        Method & Variant & Throughput$\uparrow$ & MAC\(\downarrow\) & Parameter & FID$\downarrow$ & sFID$\downarrow$ & IS$\uparrow$ & Precision$\uparrow$ & Recall$\uparrow$ \\
        \midrule

        \multirow{3}{*}{DDPM}
          & Vanilla & 0.71 & 15.56T & 113.7M & \textbf{3.51} & \underline{6.27} & \textbf{2.77} & \textbf{0.76} & \underline{0.50} \\
          & Ours (-TFS) & \underline{1.28} ($\times1.80$) & \underline{7.92T} (50.9\%) & +63.1K (0.06\%) & 4.73 & 6.98 & 2.70 & \underline{0.75} & 0.46 \\
          & \textbf{Ours} & \textbf{1.33} ($\times1.87$)& \textbf{7.87T} (50.6\%) & +102.0K (0.09\%) & \underline{3.81} & \textbf{6.00} & \textbf{2.77} & 0.74 & \textbf{0.51} \\
        \midrule

        \multirow{3}{*}{DDIM (T=50)}
          & Vanilla & 14.56 & 778.00G & 113.7M & \textbf{5.58} & \underline{8.37} & \underline{2.81} & \textbf{0.74} & \underline{0.46} \\
          & Ours (-TFS) & \underline{26.05} ($\times1.79$) & \underline{395.88G} (51.1\%) & +63.1K (0.06\%) & 7.43 & 9.07 & 2.75 & \underline{0.71} & 0.42 \\
          & \textbf{Ours} & \textbf{26.83} ($\times1.84$) & \textbf{393.63G} (50.6\%) & +102.0K (0.09\%) & \underline{6.52} & \textbf{7.82} & \textbf{2.84} & 0.68 & \textbf{0.48} \\
        \midrule

        \multirow{3}{*}{FM}
          & Vanilla & 2.12 & 589.55G & 169.1M & \textbf{2.80} & \underline{5.35} & \textbf{3.23} & \underline{0.67} & \textbf{0.60} \\
          & Ours (-TFS) & \underline{4.06} ($\times1.92$) & \underline{297.24G} (50.4\%) & +185.9K (0.11\%) & 3.67 & 5.67 & 3.17 & \textbf{0.68} & 0.56 \\
          & \textbf{Ours} & \textbf{4.08} ($\times1.92$) & \textbf{296.49G} (50.3\%) & +323.4K (0.19\%) & \underline{2.98} & \textbf{5.20} & \underline{3.22} & 0.66 & \underline{0.59} \\
        \bottomrule
    \end{tabular}
    \end{adjustbox}
    \vspace{-0.3cm}
\end{table*}

\paragraph{Noisy targets and trajectory-dependent allocation.}
The clean-manifold argument concerns the $Kd$ intrinsic coordinates of a clean tuple. With nonzero noise scale, independent Gaussian noise introduces variation in all $D$ ambient directions of each sample, giving the noisy tuple full support in $\mathbb R^{KD}$. Thus, for $K>1$, continuous fusion into $\mathbb R^D$ cannot distinguish every noisy tuple (Appendix~\ref{sec:target_details}). We therefore return to target agreement within fibers, quantified by Eq.~\ref{eq:main_ambiguity}. For velocity targets, $m_t$ is the conditional mean velocity of each sample. Small variation of this field within a fiber is precisely the vector-field compatibility needed for shared prediction.

To understand how noise affects this agreement, we must account for the prediction target. Write the forward-noised tuple as $X_t=a_tX_0+\sigma_t\epsilon$, where $a_t$ scales the clean signal, $\sigma_t>0$ scales the noise, and $\epsilon$ is the Gaussian noise tuple. For noise prediction, rearranging $\epsilon=(X_t-a_tX_0)/\sigma_t$ and conditioning on $X_t$ yields
\begin{equation}
m_t^\epsilon(X_t)=\frac{X_t-a_t\mathbb E_t[X_0\mid X_t]}{\sigma_t}.
\label{eq:main_noise_target}
\end{equation}
Here $m_t^\epsilon$ denotes the optimal noise prediction. Even at the pure-noise limit $a_t=0,\sigma_t=1$, it equals $X_t$, whereas clean-state prediction uses the posterior mean of $X_0$. Thus noise level and target parameterization together determine the information required by fusion.

Choosing when to share more computation also requires knowing how prediction errors affect sampling. For a velocity model, let $t'$ be the next sampling time, $h_t=t'-t$, $\widehat v_t=G_t(F_K(X_t))$, and $v_t^*=m_t(X_t)$. Subtracting the Euler updates from the same state, $\widehat X_{t'}=X_t+h_t\widehat v_t$ and $X^*_{t'}=X_t+h_tv_t^*$, gives
\begin{equation}
\|\widehat X_{t'}-X^*_{t'}\|=|h_t|\,\|\widehat v_t-v_t^*\|.
\label{eq:main_update_error}
\end{equation}
For a fixed velocity discrepancy, a smaller interval reduces its contribution to that update. This motivates larger fusion factors at smaller Flow Matching intervals. In DDPM/DDIM, coarse-to-fine generation motivates sharing more computation early and restoring sample-specific capacity during refinement. TFS applies these design principles, and its schedule ablations evaluate the resulting quality--throughput trade-off.

\section{Experiments}

\begin{figure*}[t]
    \centering
    \begin{minipage}{\linewidth}
        \centering
        \begin{minipage}{0.49\linewidth}
            \centering
            \begin{adjustbox}{max width=\linewidth}
                \begin{tabular}{l|l|ccccc}
                    \toprule
                    \multicolumn{7}{c}{\textbf{CIFAR-10}} \\
                    \cmidrule(lr){1-7}
                    Throughput & Variant & FID$\downarrow$ & sFID$\downarrow$ & IS$\uparrow$ & Prec.$\uparrow$ & Rec.$\uparrow$ \\
                    \midrule
                    \multirow{2}{*}{66.80 img/s}
                      & Vanilla        & 10.36 &  8.16 & 8.54 & \textbf{0.63} & 0.53 \\
                      & Ours            & \textbf{10.35} & \textbf{6.29} & \textbf{8.58} & 0.60 & \textbf{0.57} \\
                    \midrule
                    \multirow{2}{*}{156.56 img/s}
                      & Vanilla        & 17.55 & 12.66 & 8.15 & \textbf{0.59} & 0.46 \\
                      & Ours            & \textbf{14.30} & \textbf{8.76} & \textbf{8.32} & 0.58 & \textbf{0.53} \\
                    \midrule
                    \multirow{2}{*}{317.89 img/s}
                      & Vanilla        & 43.59 & 26.38 & 6.83 & 0.50 & 0.30 \\
                      & Ours            & \textbf{22.99} & \textbf{13.76} & \textbf{7.84} & \textbf{0.55} & \textbf{0.46} \\
                    \midrule
                    \multirow{2}{*}{565.03 img/s}
                      & Vanilla        & 99.09 & 51.83 & 4.80 & 0.38 & 0.15 \\
                      & Ours            & \textbf{50.07} & \textbf{28.11} & \textbf{6.48} & \textbf{0.48} & \textbf{0.31} \\
                    \bottomrule
                \end{tabular}
            \end{adjustbox}
        \end{minipage}\hfill
        \begin{minipage}{0.49\linewidth}
            \centering
            \begin{adjustbox}{max width=\linewidth}
                \begin{tabular}{l|l|ccccc}
                    \toprule
                    \multicolumn{7}{c}{\textbf{CelebA}} \\
                    \cmidrule(lr){1-7}
                    Throughput & Variant & FID$\downarrow$ & sFID$\downarrow$ & IS$\uparrow$ & Prec.$\uparrow$ & Rec.$\uparrow$ \\
                    \midrule
                    \multirow{2}{*}{26.83 img/s}
                      & Vanilla        &  7.12 & 10.68 & 2.77 & \textbf{0.73} & 0.43 \\
                      & Ours            & \textbf{6.52} & \textbf{7.82} & \textbf{2.84} & 0.68 & \textbf{0.48} \\
                    \midrule
                    \multirow{2}{*}{66.90 img/s}
                      & Vanilla        & 12.33 & 19.01 & 2.59 & \textbf{0.73} & 0.30 \\
                      & Ours            & \textbf{8.90} & \textbf{11.53} & \textbf{2.72} & 0.69 & \textbf{0.40} \\
                    \midrule
                    \multirow{2}{*}{127.92 img/s}
                      & Vanilla        & 23.06 & 31.96 & 2.37 & 0.68 & 0.15 \\
                      & Ours            & \textbf{14.11} & \textbf{18.63} & \textbf{2.53} & \textbf{0.69} & \textbf{0.28} \\
                    \midrule
                    \multirow{2}{*}{228.77 img/s}
                      & Vanilla        & 65.15 & 68.64 & 2.13 & 0.40 & 0.01 \\
                      & Ours            & \textbf{31.24} & \textbf{36.63} & \textbf{2.26} & \textbf{0.60} & \textbf{0.10} \\
                    \bottomrule
                \end{tabular}
            \end{adjustbox}
        \end{minipage}
    \end{minipage}

    \includegraphics[width=1.0\textwidth]{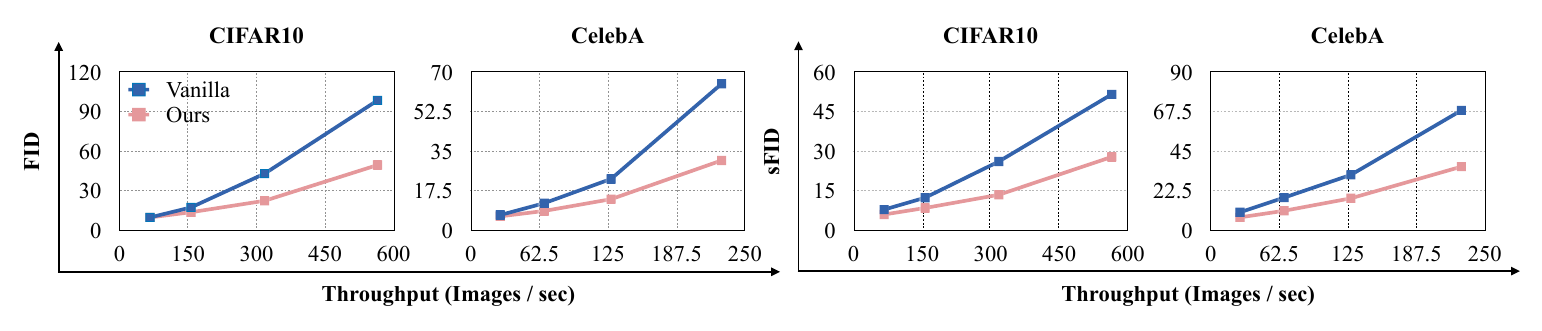}
\vspace{-0.5cm}
    \caption{Comparison of throughput vs. quality and diversity on CIFAR-10 (left) and CelebA (right). The table (top) reports FID, sFID, IS, Precision, and Recall across throughput levels for Vanilla and Ours). The plots (bottom) visualize FID (right) and sFID (left) for DDIM, highlighting how Ours (red) compares against Vanilla DDIM (blue).}
    \label{fig:throughput_quality_tradeoff}
\end{figure*}

\paragraph{\textbf{Diffusion Models and Baselines.}}

Our goal is to validate \algname{} as a general computational restructuring principle that operates across diverse generative settings, rather than a technique specialized to a particular pretrained foundation model stack. We therefore evaluate \algname{} across major diffusion model families spanning different denoising spaces, backbone architectures, and conditional generation settings. For pixel-space U-Nets, we apply \textit{\algacronym{}-UNet} to DDPM~\cite{ho2020denoising} on CIFAR-10 and CelebA. For pixel-space transformers and conditional generation, we evaluate \textit{\algacronym{}-DiT} on class-conditional DiT~\cite{peebles2023scalable} on CIFAR-10. For latent-space autoencoders, we apply \textit{\algacronym{}-AE} to LDM~\cite{rombach2022high} on CelebA-HQ. Finally, for latent-space transformers and conditional generation, we evaluate \textit{\algacronym{}-DiT} on ImageNet using DiT~\cite{peebles2023scalable}. These experiments demonstrate that \algname{} generalizes across diffusion model families rather than specific architectures, scaling across resolutions, denoising spaces, and conditioning regimes.

\paragraph{\textbf{Experimental settings.}}
All throughput measurements are conducted using batched inference for vanilla baselines and \algname{} under identical inference environments, including sufficiently large batch configurations, scheduler settings, and hardware setup, to ensure fair comparison. Throughput is reported using end-to-end wall-clock generation time, and computational overhead is evaluated using parameter count and multiply–accumulate operations (MACs). Unless otherwise specified, all \algname{} results use fusion factors up to K=4, and all models are trained from scratch without finetuning or freezing. Additional details and ablations appear in the appendix.

\subsection{Application to Baseline Models}
\label{sec:\algacronym{}_perform_standard}

We first evaluate \algname{} on baseline diffusion models, where it achieves nearly 2$\times$ throughput while maintaining sample quality close to the baseline without materially increasing model complexity(Tab.~\ref{tab:core}). Qualitative results in the Appendix confirm preserved perceptual fidelity and diversity. We observe similar behavior in transformer-based diffusion models. Interestingly, mixing different class conditions within fused groups is more stable than grouping identical classes, as verified by additional experiments in the Appendix. This supports the interpretation that conditional diversity provides cues for slot separation (Appendix~\ref{sec:target_details}).

We extend \algname{} to latent diffusion in two ways. On CelebA-HQ, we apply \textit{\algacronym{}-AE} to LDM(Tab.~\ref{tab:ldm}). We also implement \algacronym{}-UNet directly on the LDM backbone, with results provided in the Appendix. Empirically, \algacronym{}-AE exhibits smaller quality degradation than \algacronym{}-UNet at the same fusion level. This comparison highlights the role of where the shared bottleneck is introduced. \algacronym{}-AE learns a representation for joint reconstruction, whereas \algacronym{}-UNet shares the prediction computation at each denoising step. We further evaluate \textit{\algacronym{}-DiT} on ImageNet for large-scale class-conditional generation. These results show strong quality across latent U-Net and transformer backbones in both settings.

\begin{table*}[t]
\centering
\caption{Combination of \algname{} w.o.\ TFS and \algname{} with DeepCache on CIFAR-10 and CelebA. \algname{} w.o.\ TFS uses a fixed fusion factor with \(K=2\), while \algname{} dynamically allocates the fusion factor via TFS. MAC reduction ratios and throughput speedups are shown in parentheses. \textsuperscript{*}DDIM steps are reduced to match the same throughput setting.}
\vspace{1mm}
\label{tab:deepcache}

\setlength{\tabcolsep}{3.5pt}
\renewcommand{\arraystretch}{1.08}

\begin{adjustbox}{max width=0.8\linewidth}
\begin{tabular}{l|l|ll|ccccc}
\toprule
\multicolumn{9}{c}{\textbf{CIFAR-10}} \\
\midrule
\multicolumn{1}{c}{Setting} &
\multicolumn{1}{c}{Variant} &
\multicolumn{1}{c}{Throughput$\uparrow$} &
\multicolumn{1}{c}{MAC$\downarrow$ (G)} &
\multicolumn{1}{c}{FID$\downarrow$} &
\multicolumn{1}{c}{sFID$\downarrow$} &
\multicolumn{1}{c}{IS$\uparrow$} &
\multicolumn{1}{c}{Prec$\uparrow$} &
\multicolumn{1}{c}{Rec$\uparrow$} \\
\midrule

\multirow[c]{3}{*}{DDIM}
 & Vanilla & 17.70 & 606.88 & 6.30 & 5.65 & 8.88 & 0.65 & 0.56 \\
 & +DeepCache & 28.05 ($\times$1.59) & 390.18 (64.3\%) & 6.90 & 6.21 & 8.89 & 0.64 & 0.56 \\
 & \; +Step-Reduction & 49.02\textsuperscript{*} ($\times$2.77) & 207.88 (34.3\%) & 10.12 & 7.40 & 8.56 & 0.63 & 0.53 \\
\midrule

\multirow[c]{2}{*}{Ours (-TFS)}
 & Vanilla & 33.48 ($\times$1.89) & 306.88 (50.6\%) & 11.21 & 6.12 & 8.30 & 0.59 & 0.57 \\
 & +DeepCache & 51.19 ($\times$2.89) & 198.48 (32.7\%) & 11.24 & 6.48 & 8.29 & 0.59 & 0.56 \\
\midrule

\multirow[c]{2}{*}{\textbf{Ours}}
 & Vanilla & 32.90 ($\times$1.86) & 307.76 (50.7\%) & 8.65 & 5.46 & 8.68 & 0.61 & 0.59 \\
 & +DeepCache & 51.98\textsuperscript{*} ($\times$2.94) & 197.38 (32.5\%) & 9.42 & 6.37 & 8.62 & 0.61 & 0.57 \\

\midrule
\multicolumn{9}{c}{\textbf{CelebA}} \\
\midrule

\multicolumn{1}{c}{Setting} &
\multicolumn{1}{c}{Variant} &
\multicolumn{1}{c}{Throughput$\uparrow$} &
\multicolumn{1}{c}{MAC$\downarrow$ (G)} &
\multicolumn{1}{c}{FID$\downarrow$} &
\multicolumn{1}{c}{sFID$\downarrow$} &
\multicolumn{1}{c}{IS$\uparrow$} &
\multicolumn{1}{c}{Prec$\uparrow$} &
\multicolumn{1}{c}{Rec$\uparrow$} \\
\midrule

\multirow[c]{3}{*}{DDIM}
 & Vanilla & 7.20 & 1556.01 & 4.17 & 6.70 & 2.82 & 0.74 & 0.50 \\
 & +DeepCache & 9.81 ($\times$1.36) & 1075.81 (69.1\%) & 6.07 & 9.10 & 2.79 & 0.72 & 0.45 \\
 & \; +Step-Reduction & 16.83\textsuperscript{*} ($\times$2.34) & 572.58 (36.8\%) & 8.22 & 11.38 & 2.73 & 0.70 & 0.40 \\
\midrule

\multirow[c]{2}{*}{Ours (-TFS)}
 & Vanilla & 13.25 ($\times$1.84) & 791.76 (50.9\%) & 6.17 & 7.71 & 2.76 & 0.71 & 0.45 \\
 & +DeepCache & 16.96 ($\times$2.36) & 551.66 (35.5\%) & 8.43 & 10.50 & 2.71 & 0.69 & 0.40 \\
\midrule

\multirow[c]{2}{*}{\textbf{Ours}}
 & Vanilla & 13.41 ($\times$1.86) & 787.26 (50.6\%) & 5.54 & 6.91 & 2.85 & 0.69 & 0.51 \\
 & +DeepCache & 17.65\textsuperscript{*} ($\times$2.45) & 547.18 (35.2\%) & 7.07 & 8.69 & 2.79 & 0.68 & 0.46 \\
\bottomrule

\end{tabular}
\end{adjustbox}
\end{table*}

\begin{table}[t]
    \centering
    \caption{Performance of \algname{} in latent diffusion settings. \algacronym{}-AE is evaluated on CelebA-HQ using LDM, while \algacronym{}-DiT is evaluated on ImageNet using DiT-S/2 with fusion factors \(K=2,4\). For comparison with throughput gains from larger patch sizes, we also report DiT-S/4 results.}
    \label{tab:ldm}
    \begin{adjustbox}{max width=0.9\linewidth}
        \begin{tabular}{l | l l | c | c c c c c}
            \toprule
            \multicolumn{9}{c}{\textbf{CelebA-HQ}} \\
            \midrule
           \multicolumn{1}{c}{Method} & \multicolumn{1}{c}{Throughput$\uparrow$} & \multicolumn{1}{c}{MAC$\downarrow$ (G)} & \multicolumn{1}{c}{Parameter$\downarrow$} & \multicolumn{1}{c}{FID$\downarrow$} & \multicolumn{1}{c}{sFID$\downarrow$} & \multicolumn{1}{c}{IS$\uparrow$} & \multicolumn{1}{c}{Prec$\uparrow$} & \multicolumn{1}{c}{Rec$\uparrow$} \\
            \midrule
            LDM (baseline) & 0.54 & 96.38T & 362.3M & 11.58 & 12.68 & 3.12 & 0.68 & 0.34 \\
            \algacronym{}-AE (K=2) & 1.08\,($\times2.00$) & 48.19T (50.0\%) & +159.8K (0.04\%) & 12.90 & 13.74 & 2.96 & 0.66 & 0.33 \\
            \algacronym{}-AE (K=4) & 2.16\,($\times4.00$) & 24.09T (25.0\%) & +319.6K (0.08\%) & 14.57 & 16.79 & 2.78 & 0.69 & 0.24 \\
            \midrule
            \multicolumn{9}{c}{\textbf{ImageNet}} \\
            \midrule
            \multicolumn{1}{c}{Method} & \multicolumn{1}{c}{Throughput$\uparrow$} & \multicolumn{1}{c}{MAC$\downarrow$ (G)} & \multicolumn{1}{c}{Parameter$\downarrow$} & \multicolumn{1}{c}{FID$\downarrow$} & \multicolumn{1}{c}{sFID$\downarrow$} & \multicolumn{1}{c}{IS$\uparrow$} & \multicolumn{1}{c}{Prec$\uparrow$} & \multicolumn{1}{c}{Rec$\uparrow$} \\
            \midrule
            DiT-S/2 (baseline) & 3.04 & 1067.50G & 32.9M & 71.85 & 12.29 & 18.65 & 0.35 & 0.55 \\
            Ours (-TFS) & 5.95\,($\times1.96$) & 562.50G (52.69\%) & +1.33M (4.04\%) & 78.22 & 15.38 & 16.83 & 0.34 & 0.53 \\
            \textbf{Ours} & 5.98\,($\times1.97$) & 551.95G (51.70\%) & +1.94M(5.90\%) & 74.19 & 14.85 & 19.08 & 0.34 & 0.49 \\
            \bottomrule
        \end{tabular}
    \end{adjustbox}
    \vspace{-0.3cm}
\end{table}

\subsection{Orthogonal Acceleration Beyond Existing Throughput Limits}
\label{sec:\algacronym{}_perform_step}

We extend \algname{} to step-reduction frameworks, demonstrating that per-step multi-sample acceleration is complementary to trajectory shortening. As shown in Fig.~\ref{fig:throughput_quality_tradeoff}, \algname{} maintains higher sample quality at matched throughput. In contrast, baseline methods rapidly degrade beyond their practical throughput ceiling, highlighting the advantage of trajectory-dependent fusion allocation in high-throughput regimes.

Finally, we combine \algname{} with DeepCache to demonstrate compatibility with caching-based acceleration. At matched throughput, the combined approach consistently outperforms configurations relying solely on step-reduction and caching(Tab.~\ref{tab:deepcache}). These results suggest that compressing computation along a single axis is fundamentally limited, whereas combining orthogonal acceleration strategies yields compounding efficiency gains in high-throughput settings.

\section{Analysis}
\label{sec:analysis}

\begin{figure}[t]
    \centering
    \includegraphics[width=\linewidth]{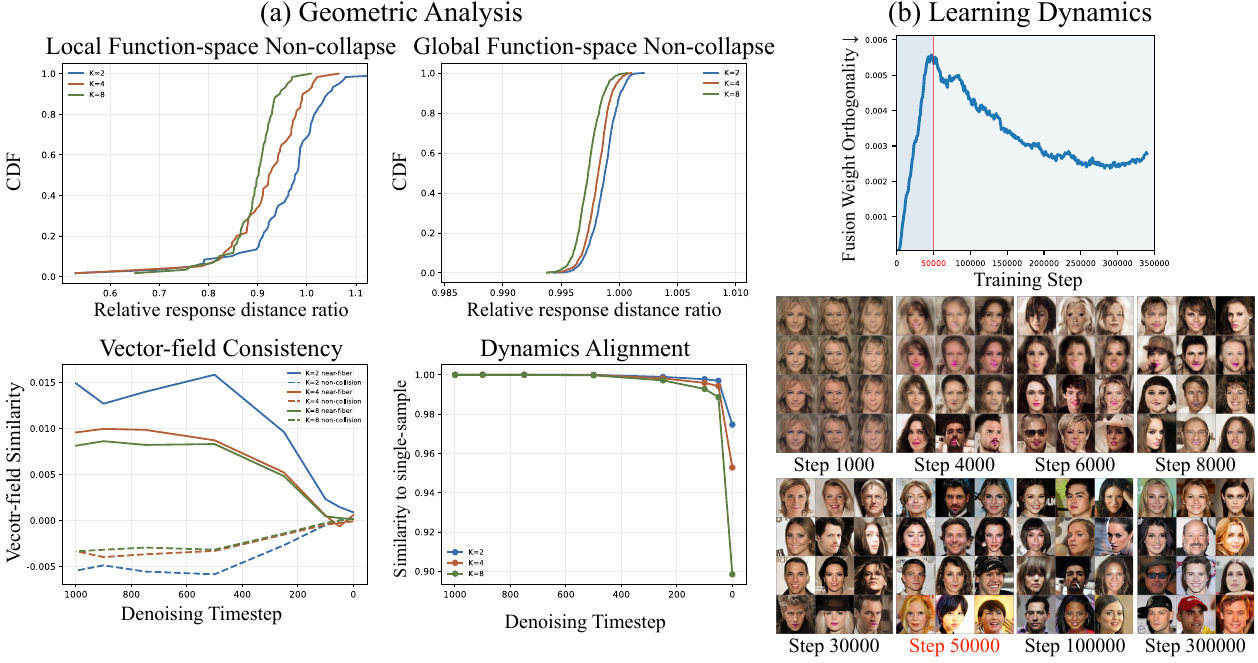}
    \caption{Analysis of the representation geometry induced by \algname{}.
    (a) Geometric diagnostics measure prediction separation using response-distance ratios, and vector-field consistency and dynamics alignment using cosine similarity.
    (b) Training dynamics of Fusion Weight Orthogonality for MF-AE with $K=4$, showing the development of slot-specific separation during training.}
    \label{fig:analysis}
\end{figure}

We analyze the geometry and dynamics induced by multiplexed generation under \algname{}. In particular, we study whether fusion preserves distinguishability, maintains consistent generative dynamics, and develops stable separation during training. To this end, we examine prediction separation, vector-field consistency, dynamics alignment, and fusion-module behavior. Additional visualizations and toy analysis on the $D \ge Kd$ condition are provided in the appendix.

\subsection{Empirical Validation of the Geometric Analysis}

To examine whether fusion retains prediction-relevant distinctions (Sec.~\ref{subsec:geometric_analysis}), we compare velocity-prediction distances with those of the single-sample model. Figure~\ref{fig:analysis}(a) reports their ratios for same-class local perturbations and random tuple pairs. Ratios near one indicate comparable prediction separation. The random-pair ratios concentrate near one, whereas local ratios shift downward as $K$ increases. Fusion thus preserves broad prediction differences while attenuating finer distinctions under stronger compression. The Appendix further examines local sensitivity.

Sharing a representation also requires compatible predictions. To examine this requirement, we compare slot-wise vector-field directions between complete tuples. Collision pairs have nearby fused representations, approximating a shared fiber, while non-collision pairs are farther apart. Near-fiber pairs exhibit higher cosine similarity over much of the trajectory, indicating a relative tendency toward more compatible field directions. To connect this agreement to generation, we also assess alignment with the reference dynamics. Alignment with the single-sample model remains close to one in early stages but declines near the end, particularly for $K=8$. These results motivate TFS, which shares more computation early and restores sample-specific capacity during final refinement. Schedule ablations evaluate the quality--throughput trade-off.

\subsection{Training Dynamics of the Fusion Module}
To further understand the fusion mechanism, we analyze the training dynamics of the fusion module through \textit{Fusion Weight Orthogonality}, defined as the average squared inner product between normalized fusion weight vectors. Figure~\ref{fig:analysis} shows that the value initially increases and then decreases after approximately $50\mathrm{k}$ steps. Early in training, different streams produce nearly identical faces, and separation emerges progressively as individual streams diverge in structure. Once the generated images begin to exhibit distinct color patterns across streams, Fusion Weight Orthogonality starts to decrease, indicating that the fusion weight vectors become increasingly orthogonal. This suggests that the fusion module gradually reorganizes its parameter space to allocate independent slot-specific directions for different samples, supporting the trajectory-wise separation discussed in Sec.~\ref{subsec:geometric_analysis}.

\section{Related Work}

\paragraph{\textbf{Diffusion Models}}
Diffusion models have become a dominant class of generative models by learning to iteratively denoise corrupted signals~\cite{sohl2015deep,ho2020denoising,song2019generative}. Pixel-space approaches such as DDPM operate directly in image space~\cite{ho2020denoising}, while Latent Diffusion Models (LDMs) perform denoising in compressed but still high-capacity latent space~\cite{rombach2022high}. Recent transformer-based variants such as DiT~\cite{peebles2023scalable} demonstrate that larger latent representations and L/XL-scale backbones improve performance but also increase inference cost. Across these formulations, diffusion models leave substantial representational slack for \algname{}, increasing the potential of fused multi-sample computation at scale.

\paragraph{\textbf{Diffusion Model Acceleration}}
The primary bottleneck of diffusion models is iterative inference, where generating a single sample requires many sequential denoising steps and repeated network evaluations. A large body of work therefore focuses on reducing the number of sampling steps. For example, DDIM reformulates the reverse process with fewer iterations~\cite{song2020denoising,nichol2021improved}, while flow-matching methods learn continuous-time vector fields that directly transport noise to data~\cite{lipman2022flow,lipman2024flow}. Another line of work reduces the cost of each sampling step through caching. Methods such as DeepCache and CacheMe reuse intermediate activations across timesteps to avoid redundant computation and lower per-step latency.~\cite{noroozi2024you,xia2023towards,yue2023resshift} However, these approaches still process each trajectory independently and therefore do not improve multi-sample throughput within a single denoising evaluation.

\paragraph{\textbf{Superposition in Deep Learning}} A line of work has explored accelerating neural networks by superposing
multiple inputs into a shared representation.
For example, the Byte Latent Transformer~\cite{bytelate} leverages sparsity to process
groups of tokens jointly, while Latent Attention~\cite{liu2024deepseek} fuses
multiple tokens into a shared latent representation for attention. Liu.al.~\cite{murahari2022datamux} extended input superposition to sentence- and token-level classification, while Menet et al.~\cite{menet2023mimonets} applied the concept to the classification of small-scale image datasets such as MNIST and CIFAR-10.
In contrast, we are the first to apply superposition to generative modeling by
fusing multiple signals within a latent trajectory, allowing one inference
process to generate multiple images without quality loss.

\section{Conclusion}
In this paper, we proposed \algname{}, a diffusion acceleration framework that improves throughput by processing multiple samples within a single denoising evaluation. Rather than relying only on trajectory shortening or per-step latency reduction, \algname{} introduces a complementary throughput axis by increasing the number of samples processed per evaluation. Our geometric analysis explains how shared representations can retain sample-wise prediction information, while trajectory-aware fusion allocation balances throughput and generation quality. Across diverse diffusion settings, \algname{} operates on top of existing acceleration strategies and enables throughput regimes difficult to achieve with them alone. These results position \algname{} as a scalable direction for future diffusion acceleration.

\paragraph{Limitations}
A limitation of \algname{} is that it currently requires full pretraining to integrate fused computation into the denoising backbone, resulting in substantial computational cost and preventing lightweight finetuning-based adaptation. In addition, we do not evaluate large-scale text-to-image foundation models, limiting our study to discussing potential scalability rather than empirically validating such settings. Finally, performance can still degrade under extreme fusion factors, suggesting remaining limits in high-throughput scaling.

\clearpage

\renewcommand{\acksection}{\section*{Acknowledgements}}
\begin{ack}
This work was supported by the Korean Government through the grants from IITP (RS-2021-II211343, RS-2022-II220320, RS-2025-25442338) and Samsung Electronics Co., Ltd (IO251219-14870-01).
\end{ack}

\medskip

{
    \small
    \bibliographystyle{plain}
    \bibliography{main}

@article{ho2020denoising,
  title={Denoising diffusion probabilistic models},
  author={Ho, Jonathan and Jain, Ajay and Abbeel, Pieter},
  journal={Advances in neural information processing systems},
  volume={33},
  pages={6840--6851},
  year={2020}
}

@inproceedings{rombach2022high,
  title={High-resolution image synthesis with latent diffusion models},
  author={Rombach, Robin and Blattmann, Andreas and Lorenz, Dominik and Esser, Patrick and Ommer, Bj{\"o}rn},
  booktitle={2022 IEEE/CVF conference on computer vision and pattern recognition (CVPR)},
  pages={10674--10685},
  year={2022},
  organization={ieee}
}

@article{lipman2022flow,
  title={Flow matching for generative modeling},
  author={Lipman, Yaron and Chen, Ricky TQ and Ben-Hamu, Heli and Nickel, Maximilian and Le, Matt},
  journal={arXiv preprint arXiv:2210.02747},
  year={2022}
}

@article{song2020denoising,
  title={Denoising diffusion implicit models},
  author={Song, Jiaming and Meng, Chenlin and Ermon, Stefano},
  journal={arXiv preprint arXiv:2010.02502},
  year={2020}
}

@article{song2023consistency,
  title={Consistency models},
  author={Song, Yang and Dhariwal, Prafulla and Chen, Mark and Sutskever, Ilya},
  journal={arXiv preprint arXiv:2303.01469},
  year={2023}
}

@inproceedings{ma2024deepcache,
  title={Deepcache: Accelerating diffusion models for free},
  author={Ma, Xinyin and Fang, Gongfan and Wang, Xinchao},
  booktitle={2024 IEEE/CVF Conference on Computer Vision and Pattern Recognition (CVPR)},
  pages={15762--15772},
  year={2024},
  organization={IEEE}
}

@article{murahari2022datamux,
  title={Datamux: Data multiplexing for neural networks},
  author={Murahari, Vishvak and Jimenez, Carlos and Yang, Runzhe and Narasimhan, Karthik},
  journal={Advances in Neural Information Processing Systems},
  volume={35},
  pages={17515--17527},
  year={2022}
}

@article{menet2023mimonets,
  title={Mimonets: Multiple-input-multiple-output neural networks exploiting computation in superposition},
  author={Menet, Nicolas and Hersche, Michael and Karunaratne, Geethan and Benini, Luca and Sebastian, Abu and Rahimi, Abbas},
  journal={Advances in Neural Information Processing Systems},
  volume={36},
  pages={39553--39565},
  year={2023}
}

@inproceedings{ronneberger2015u,
  title={U-net: Convolutional networks for biomedical image segmentation},
  author={Ronneberger, Olaf and Fischer, Philipp and Brox, Thomas},
  booktitle={International Conference on Medical image computing and computer-assisted intervention},
  pages={234--241},
  year={2015},
  organization={Springer}
}

@inproceedings{nichol2021improved,
  title={Improved denoising diffusion probabilistic models},
  author={Nichol, Alexander Quinn and Dhariwal, Prafulla},
  booktitle={International conference on machine learning},
  pages={8162--8171},
  year={2021},
  organization={PmLR}
}

@article{lipman2024flow,
  title={Flow matching guide and code},
  author={Lipman, Yaron and Havasi, Marton and Holderrieth, Peter and Shaul, Neta and Le, Matt and Karrer, Brian and Chen, Ricky TQ and Lopez-Paz, David and Ben-Hamu, Heli and Gat, Itai},
  journal={arXiv preprint arXiv:2412.06264},
  year={2024}
}

@article{nash2021generating,
  title={Generating images with sparse representations},
  author={Nash, Charlie and Menick, Jacob and Dieleman, Sander and Battaglia, Peter W},
  journal={arXiv preprint arXiv:2103.03841},
  year={2021}
}

@article{dhariwal2021diffusion,
  title={Diffusion models beat gans on image synthesis},
  author={Dhariwal, Prafulla and Nichol, Alexander},
  journal={Advances in neural information processing systems},
  volume={34},
  pages={8780--8794},
  year={2021}
}

@article{salimans2016improved,
  title={Improved techniques for training gans},
  author={Salimans, Tim and Goodfellow, Ian and Zaremba, Wojciech and Cheung, Vicki and Radford, Alec and Chen, Xi},
  journal={Advances in neural information processing systems},
  volume={29},
  year={2016}
}

@article{heusel2017gans,
  title={Gans trained by a two time-scale update rule converge to a local nash equilibrium},
  author={Heusel, Martin and Ramsauer, Hubert and Unterthiner, Thomas and Nessler, Bernhard and Hochreiter, Sepp},
  journal={Advances in neural information processing systems},
  volume={30},
  year={2017}
}

@article{kynkaanniemi2019improved,
  title={Improved precision and recall metric for assessing generative models},
  author={Kynk{\"a}{\"a}nniemi, Tuomas and Karras, Tero and Laine, Samuli and Lehtinen, Jaakko and Aila, Timo},
  journal={Advances in neural information processing systems},
  volume={32},
  year={2019}
}

@inproceedings{peebles2023scalable,
  title={Scalable diffusion models with transformers},
  author={Peebles, William and Xie, Saining},
  booktitle={2023 IEEE/CVF International Conference on Computer Vision (ICCV)},
  pages={4172--4182},
  year={2023},
  organization={IEEE}
}

@article{lu2022dpm,
  title={Dpm-solver: A fast ode solver for diffusion probabilistic model sampling in around 10 steps},
  author={Lu, Cheng and Zhou, Yuhao and Bao, Fan and Chen, Jianfei and Li, Chongxuan and Zhu, Jun},
  journal={Advances in neural information processing systems},
  volume={35},
  pages={5775--5787},
  year={2022}
}

@article{lu2025dpm,
  title={Dpm-solver++: Fast solver for guided sampling of diffusion probabilistic models},
  author={Lu, Cheng and Zhou, Yuhao and Bao, Fan and Chen, Jianfei and Li, Chongxuan and Zhu, Jun},
  journal={Machine Intelligence Research},
  volume={22},
  number={4},
  pages={730--751},
  year={2025},
  publisher={Springer}
}

@article{zhang2022fast,
  title={Fast sampling of diffusion models with exponential integrator},
  author={Zhang, Qinsheng and Chen, Yongxin},
  journal={arXiv preprint arXiv:2204.13902},
  year={2022}
}

@inproceedings{zhou2024fast,
  title={Fast ode-based sampling for diffusion models in around 5 steps},
  author={Zhou, Zhenyu and Chen, Defang and Wang, Can and Chen, Chun},
  booktitle={2024 IEEE/CVF Conference on Computer Vision and Pattern Recognition (CVPR)},
  pages={7777--7786},
  year={2024},
  organization={IEEE}
}

@article{salimans2022progressive,
  title={Progressive distillation for fast sampling of diffusion models},
  author={Salimans, Tim and Ho, Jonathan},
  journal={arXiv preprint arXiv:2202.00512},
  year={2022}
}

@article{zhou2024simple,
  title={Simple and fast distillation of diffusion models},
  author={Zhou, Zhenyu and Chen, Defang and Wang, Can and Chen, Chun and Lyu, Siwei},
  journal={Advances in Neural Information Processing Systems},
  volume={37},
  pages={40831--40860},
  year={2024}
}

@article{li2024faster,
  title={Faster diffusion: Rethinking the role of the encoder for diffusion model inference},
  author={Li, Senmao and Hu, Taihang and Van De Weijer, Joost and Khan, Fahad S and Liu, Tao and Li, Linxuan and Yang, Shiqi and Wang, Yaxing and Cheng, Ming-Ming and Yang, Jian},
  journal={Advances in neural information processing systems},
  volume={37},
  pages={85203--85240},
  year={2024}
}

@inproceedings{sohl2015deep,
  title={Deep unsupervised learning using nonequilibrium thermodynamics},
  author={Sohl-Dickstein, Jascha and Weiss, Eric and Maheswaranathan, Niru and Ganguli, Surya},
  booktitle={International conference on machine learning},
  pages={2256--2265},
  year={2015},
  organization={pmlr}
}

@article{song2019generative,
  title={Generative modeling by estimating gradients of the data distribution},
  author={Song, Yang and Ermon, Stefano},
  journal={Advances in neural information processing systems},
  volume={32},
  year={2019}
}

@article{karras2022elucidating,
  title={Elucidating the design space of diffusion-based generative models},
  author={Karras, Tero and Aittala, Miika and Aila, Timo and Laine, Samuli},
  journal={Advances in neural information processing systems},
  volume={35},
  pages={26565--26577},
  year={2022}
}

@article{sajjadi2018assessing,
  title={Assessing generative models via precision and recall. arXiv preprint},
  author={Sajjadi, MS and Bachem, O and Lucic, M and Bousquet, O and Gelly, S},
  journal={arXiv preprint arXiv:1806.00035},
  year={2018}
}

@article{olshausen1996emergence,
  title={Emergence of simple-cell receptive field properties by learning a sparse code for natural images},
  author={Olshausen, Bruno A and Field, David J},
  journal={Nature},
  volume={381},
  number={6583},
  pages={607--609},
  year={1996},
  publisher={Nature Publishing Group UK London}
}

@article{simoncelli2001natural,
  title={Natural image statistics and neural representation},
  author={Simoncelli, Eero P and Olshausen, Bruno A},
  journal={Annual review of neuroscience},
  volume={24},
  number={1},
  pages={1193--1216},
  year={2001},
  publisher={Annual Reviews 4139 El Camino Way, PO Box 10139, Palo Alto, CA 94303-0139, USA}
}

@article{zhang2017mixup,
  title={mixup: Beyond empirical risk minimization},
  author={Zhang, Hongyi and Cisse, Moustapha and Dauphin, Yann N and Lopez-Paz, David},
  journal={arXiv preprint arXiv:1710.09412},
  year={2017}
}

@inproceedings{verma2019manifold,
  title={Manifold mixup: Better representations by interpolating hidden states},
  author={Verma, Vikas and Lamb, Alex and Beckham, Christopher and Najafi, Amir and Mitliagkas, Ioannis and Lopez-Paz, David and Bengio, Yoshua},
  booktitle={International conference on machine learning},
  pages={6438--6447},
  year={2019},
  organization={PMLR}
}

@inproceedings{noroozi2024you,
  title={You only need one step: Fast super-resolution with stable diffusion via scale distillation},
  author={Noroozi, Mehdi and Hadji, Isma and Martinez, Brais and Bulat, Adrian and Tzimiropoulos, Georgios},
  booktitle={European Conference on Computer Vision},
  pages={145--161},
  year={2024},
  organization={Springer}
}

@misc{xia2023towards,
      title={Towards More Accurate Diffusion Model Acceleration with A Timestep Tuner}, 
      author={Mengfei Xia and Yujun Shen and Changsong Lei and Yu Zhou and Ran Yi and Deli Zhao and Wenping Wang and Yong-Jin Liu},
      year={2025},
      eprint={2310.09469},
      archivePrefix={arXiv},
      primaryClass={cs.CV},
      url={https://arxiv.org/abs/2310.09469}, 
}

@inproceedings{yue2023resshift,
  author = {Zongsheng Yue and Jianyi Wang and Chen Change Loy},
  title     = {ResShift: Efficient Diffusion Model for Image Super-resolution by Residual Shifting},
  booktitle = {Advances in Neural Information Processing Systems (NeurIPS)},
  year      = {2023},
}

@misc{yang2023denoising,
      title={Denoising Diffusion Step-aware Models}, 
      author={Shuai Yang and Yukang Chen and Luozhou Wang and Shu Liu and Yingcong Chen},
      year={2024},
      eprint={2310.03337},
      archivePrefix={arXiv},
      primaryClass={cs.CV},
      url={https://arxiv.org/abs/2310.03337}, 
}

@article{carlsson2009topology,
  title={Topology and data},
  author={Carlsson, Gunnar},
  journal={Bulletin of the American mathematical society},
  volume={46},
  number={2},
  pages={255--308},
  year={2009}
}

@article{chapelle2006semi,
  title={Semi-supervised learning (chapelle, o. et al., eds.; 2006)[book reviews]},
  author={Chapelle, Olivier and Scholkopf, Bernhard and Zien, Alexander},
  journal={IEEE Transactions on Neural Networks},
  volume={20},
  number={3},
  pages={542--542},
  year={2009},
  publisher={IEEE}
}

@article{liu2024deepseek,
  title={Deepseek-v3 technical report},
  author={Liu, Aixin and Feng, Bei and Xue, Bing and Wang, Bingxuan and Wu, Bochao and Lu, Chengda and Zhao, Chenggang and Deng, Chengqi and Zhang, Chenyu and Ruan, Chong and others},
  journal={arXiv preprint arXiv:2412.19437},
  year={2024}
}

@inproceedings{bytelate,
  title={Byte latent transformer: Patches scale better than tokens},
  author={Pagnoni, Artidoro and Pasunuru, Ramakanth and Rodriguez, Pedro and Nguyen, John and Muller, Benjamin and Li, Margaret and Zhou, Chunting and Yu, Lili and Weston, Jason E and Zettlemoyer, Luke and others},
  booktitle={Proceedings of the 63rd Annual Meeting of the Association for Computational Linguistics (Volume 1: Long Papers)},
  pages={9238--9258},
  year={2025}
}
}

\appendix

\clearpage
\appendix
\makeatletter
\setlength{\@fptop}{0pt}
\makeatother

\FloatBarrier
\section{Appendix Overview}

This appendix follows the main paper from implementation to analysis and evaluation. We first provide training and inference configurations, followed by derivations supporting the geometric analysis. We then examine clean-manifold capacity, local sensitivity, interpolation, and feature-space coverage empirically. Next come extensions through larger fusion factors, TFS, DeepCache, and latent diffusion, followed by conditional grouping and kernel-size ablations. We conclude with throughput-oriented baseline comparisons and qualitative results organized by sampler, throughput, and latent fusion design.

\FloatBarrier
\section{Implementation Details}
\label{sec:implementaion}

We describe here the training and inference configurations and metric informations used across all datasets and model families. 

\paragraph{\textbf{Metrics}}
We assess perceptual quality and distributional coverage along five complementary axes. First, Inception Score (IS) \cite{salimans2016improved} and Fréchet Inception Distance (FID) \cite{heusel2017gans} to jointly capture sample diversity and global visual fidelity. Second, we compute the spatial FID (sFID), which measures the Fréchet distance over intermediate feature maps that preserve spatial structure, providing a more sensitive evaluation of local detail and geometric fidelity \cite{nash2021generating,dhariwal2021diffusion}. We additionally report precision and recall~\cite{kynkaanniemi2019improved,sajjadi2018assessing} to quantify manifold fidelity and coverage under fusion. Moreover, all Macs, Throughputs are measured by a single NVIDIA A6000 GPU with batched inference of batch 128.

\paragraph{\textbf{Configurations}}
For CIFAR-10 DDPM, we follow the protocol of Ho et al.~\cite{ho2020denoising}, while CelebA DDPM and the DDIM samplers adopt the settings of Song et al.~\cite{song2020denoising}. For Flow Matching on CIFAR-10, we replicate Lipman et al.~\cite{lipman2024flow}. For CelebA, no official baseline exists, so we define its configuration consistently with the rest of our experiments.
For latent diffusion models, we follow the training and sampling setup of Rombach et al.~\cite{rombach2022high}. For transformer-based diffusion models, we adopt the DiT configuration proposed by Peebles and Xie~\cite{peebles2023scalable} for ImageNet.

Across all experiments, models are trained with a fixed fusion factor $K$. Each model is trained independently for its corresponding $K$, and no mixture of fusion factors is used during training. The index embeddings used for demultiplexing are therefore defined only for the corresponding $K$ and remain fixed throughout training and inference.
We do not introduce additional regularization or curriculum strategies. Separation emerges from the architecture and training objective.

For Trajectory-Fusion Scheduling(TFS), we combine models trained with different fusion factors during inference. The denoising trajectory is partitioned into intervals and each interval uses the model trained with the corresponding $K$. This allows large fusion factors in early timesteps for higher throughput while gradually reducing $K$ later to preserve generation quality.

For the denoiser-level variants, TFS switches among independently trained fixed-$K$ predictors while the sampler retains the individual sample states. Each selected predictor supplies slot-wise update targets as in Eq.~\ref{eq:target_sampler}. Changing $K$ therefore changes the amount of shared prediction computation while preserving the sample-wise sampling interface. We use trajectory stage for DDPM/DDIM and scheduler intervals for Flow Matching to allocate this computation. The latter choice follows the step-size scaling in Eq.~\ref{eq:euler_prediction_error}. Schedule ablations measure the resulting quality--throughput trade-off. \algacronym{}-AE uses a fixed $K$ throughout shared-latent generation and unfuses after decoding.

Across all cases, the same learning rate schedule, noise schedule, and sampling procedure are maintained for each fixed-$K$ model. Evaluation is performed on 50k generated samples using the implementation of Dhariwal et al.~\cite{dhariwal2021diffusion}, ensuring that any performance differences can be attributed solely to the \algname{} modules. The detailed configuration is provided in Table~\ref{tab:ImplementationDetails}.

\begin{table}[!htbp]
    \centering
    \caption{Training and inference configurations of DDPM, DDIM, and Flow Matching on CIFAR-10 and CelebA, including model architecture parameters, diffusion sampling schedules, and optimization hyperparameters.}
    \vspace{1mm}
    \begin{adjustbox}{max width=\linewidth}
        \begin{tabular}{lcccccc}
            \toprule
            & \multicolumn{3}{c}{\textbf{CIFAR-10}} & \multicolumn{3}{c}{\textbf{CelebA}} \\
              \cmidrule(lr){2-4}\cmidrule(lr){5-7}
            & DDPM & DDIM & Flow Matching & DDPM & DDIM & Flow Matching \\
            \midrule
                Channels             & 128 & 128 & 128                 & 128 & 128 & 128 \\
                Channel Multiplier   & 1,2,2,2 & 1,2,2,2 & 2,2,2           & 1,1,2,2,4,4 & 1,1,2,2,4,4 & 2,2,4,4 \\
                Number of Heads      & 1 & 1 & 1                     & 16,16,32,32,64,64 & 16,16,32,32,64,64 & 1 \\
                Activation Function  & SiLU & SiLU & SiLU              & SiLU & SiLU & SiLU \\
                Diffusion Steps      & 1000 & 50 & 50                & 1000 & 50 & 50 \\
                Noise Schedule       & linear & linear & –            & linear & linear & – \\
                Beta Start           & 0.0001 & 0.0001 & –          & 0.0001 & 0.0001 & – \\
                Beta End             & 0.02 & 0.02 & –               & 0.02 & 0.02 & – \\
                Dropout              & 0.1 & 0.1 & 0.3               & 0.0 & 0.0 & 0.3 \\
                Learning Rate        & 2e-4 & 2e-4 & 1e-4            & 2e-4 & 2e-4 & 1e-4 \\
                Batch Size           & 128 & 128 & 64               & 32 & 32 & 64 \\
            \bottomrule
        \end{tabular}
    \end{adjustbox}
    \label{tab:ImplementationDetails}
\end{table}

\FloatBarrier
\section{Derivations for the Geometric Analysis}
\label{sec:target_details}
We analyze fusion through the information required for prediction. The central principle is that complete tuples mapped to the same fused representation should require compatible prediction targets. Clean-manifold capacity, fiber-wise consistency at noisy timesteps, and trajectory-dependent fusion allocation are different aspects of this principle. The timestep-wise analysis concerns \algacronym{}-UNet and \algacronym{}-DiT. The clean-reconstruction argument also describes the representation bottleneck of \algacronym{}-AE. The risk decomposition establishes the prediction criterion. The geometric arguments explain how representation capacity and separation support it. The experiments then examine these mechanisms in learned models.

\paragraph{General targets and harmful fusion collisions.}
At timestep $t$, let $X_t=(x_t^1,\ldots,x_t^K)$ denote the input tuple and $Y_t=(y_t^1,\ldots,y_t^K)$ its training targets. A target may be noise, velocity, a score-related quantity, a flow-matching target, or a clean state. Using uppercase letters for the complete tuples in this analysis, the computation in Eq.~\ref{eq:dual_approx} becomes
\begin{equation}
Z_t=F_K(X_t),\qquad \widehat Y_t=G_t(Z_t),\qquad
m_t(X_t):=\mathbb E[Y_t\mid X_t,t].
\label{eq:target_prediction}
\end{equation}
Here $G_t$ comprises the backbone, index-conditioned unfusion, and output head. Its dependence on $K$ is implicit. The sampler combines this prediction with the individual states as described below. Any additional conditioning available to the predictor is held fixed or included in the conditioning of the expectations below. The relevant requirement is task-relevant target identifiability. Ideally, $m_t(X_t)=g_t(F_K(X_t))$ almost surely for some fused-space function $g_t$. Distinct tuples can therefore share a representation without loss when their optimal targets agree. For update targets, the sampler already receives the current states, giving
\begin{equation}
\widehat X_{t'}=S_{t\to t'}\!\left(X_t,G_t(F_K(X_t))\right),
\label{eq:target_sampler}
\end{equation}
where $t'$ is the next sampling time and $S_{t\to t'}$ denotes the update (with sampler randomness suppressed). The computation therefore has two complementary roles. The predictor $G_t$ predicts the sample-wise targets from the shared representation, and $S_{t\to t'}$ applies them to the individual states. The target parameterization determines which information the fused branch must retain.

\paragraph{Conditional means and the risk identities.}
We make explicit the steps leading to Eqs.~\ref{eq:main_ambiguity} and~\ref{eq:main_risk}. Fix $t$, the available conditioning, and a deterministic measurable fusion map $F_K$. Assume $\mathbb E_t\|Y_t\|^2<\infty$ and consider square-integrable predictors. Here $\mathbb E_t$ uses the same fixed conditioning as in the main text. Write $m=m_t(X_t)$ and let $q=q(X_t)$ be any full-input predictor. Since $\mathbb E_t[Y_t-m\mid X_t]=0$, iterated expectation gives $\mathbb E_t\langle Y_t-m,m-q\rangle=0$, where $\langle\cdot,\cdot\rangle$ is the Euclidean inner product. Expanding $Y_t-q=(Y_t-m)+(m-q)$ therefore yields
\begin{equation}
\mathbb E_t\|Y_t-q\|^2
=\mathbb E_t\|Y_t-m\|^2+\mathbb E_t\|m-q\|^2.
\label{eq:conditional_projection}
\end{equation}
The second term is nonnegative and vanishes at $q=m$. This proves that the conditional mean is the optimal full-input prediction, uniquely up to almost-sure equality. Applying the same argument to predictors observing $Z_t$ gives the optimum $\mathbb E_t[Y_t\mid Z_t]$. Because $Z_t$ is a function of $X_t$, iterated expectation identifies this optimum as $g_t^*(Z_t)=\mathbb E_t[m_t(X_t)\mid Z_t]$.

Under squared loss, define the population risks
\begin{equation}
\begin{aligned}
R_{\mathrm{full},t}&=\inf_q\mathbb E[\|Y_t-q(X_t)\|^2\mid t],\\
R_{F,t}&=\inf_g\mathbb E[\|Y_t-g(F_K(X_t))\|^2\mid t],
\end{aligned}
\label{eq:fusion_risks}
\end{equation}
where the infima range over all square-integrable predictors with the indicated inputs, for a fixed fusion map. The optimal fused predictor is $g_t^*(z)=\mathbb E[m_t(X_t)\mid F_K(X_t)=z,t]$, giving
\begin{equation}
\begin{aligned}
\Delta_t(K)&:=R_{F,t}-R_{\mathrm{full},t}\\
&=\mathbb E\!\left[\left\|m_t(X_t)-\mathbb E[m_t(X_t)\mid F_K(X_t),t]\right\|^2\mid t\right].
\end{aligned}
\label{eq:fusion_ambiguity}
\end{equation}
This identity applies to any square-integrable target $Y_t$. The excess risk is the expected conditional variance of the optimal target within each fiber. It vanishes exactly when $m_t(X_t)$ is determined by $Z_t$ almost surely. Thus, overlap between input representations is compatible with accurate prediction when the corresponding targets agree. For the finite learned predictor, let $R_{\theta,F,t}=\mathbb E[\|Y_t-G_t(F_K(X_t))\|^2\mid t]$. Its excess risk decomposes as
\begin{equation}
\begin{aligned}
R_{\theta,F,t}-R_{\mathrm{full},t}
&=\underbrace{\mathbb E[\|m_t(X_t)-g_t^*(Z_t)\|^2\mid t]}_{\text{irreducible fusion ambiguity }\Delta_t(K)}\\
&\quad+\underbrace{\mathbb E[\|g_t^*(Z_t)-G_t(Z_t)\|^2\mid t]}_{\text{approximation and optimization error}}.
\end{aligned}
\label{eq:fusion_risk_decomposition}
\end{equation}
To verify the decomposition directly, abbreviate $g^*=g_t^*(Z_t)$ and $G=G_t(Z_t)$ and split the error into three residuals,
\begin{equation}
Y_t-G=\underbrace{Y_t-m}_{A}+\underbrace{m-g^*}_{B}+\underbrace{g^*-G}_{C}.
\label{eq:residual_decomposition}
\end{equation}
The residual $A$ has zero conditional mean given $X_t$, and $B$ has zero conditional mean given $Z_t$. Since $B$ and $C$ are functions of $X_t$, and $C$ is also a function of $Z_t$, their cross terms vanish,
\[
\begin{aligned}
\mathbb E_t\langle A,B\rangle
&=\mathbb E_t\langle\mathbb E_t[A\mid X_t],B\rangle=0,\\
\mathbb E_t\langle A,C\rangle
&=\mathbb E_t\langle\mathbb E_t[A\mid X_t],C\rangle=0,\\
\mathbb E_t\langle B,C\rangle
&=\mathbb E_t\langle\mathbb E_t[B\mid Z_t],C\rangle=0.
\end{aligned}
\]
Consequently, $R_{\theta,F,t}=\mathbb E_t\|A\|^2+\mathbb E_t\|B\|^2+\mathbb E_t\|C\|^2$. The first term is $R_{\mathrm{full},t}$. Setting $G=g^*$ makes $C=0$ and gives $R_{F,t}-R_{\mathrm{full},t}=\mathbb E_t\|B\|^2$, proving Eq.~\ref{eq:main_ambiguity}. Restoring the actual $G$ adds $\mathbb E_t\|C\|^2$, proving Eq.~\ref{eq:main_risk}. Zero ambiguity is equivalent to $m=g^*$ almost surely, so the information criterion concerns fibers represented in the data distribution. This decomposition explains the learning pressure on the fusion modules. For a fixed $F_K$, the predictor approaches the conditional mean of the retained targets. Updating $F_K$ also changes the fibers and hence the ambiguity term. End-to-end training can therefore improve both the organization of target information and its recovery, linking the regression objective to the emergence of slot separation.

\paragraph{Clean-state prediction as an interpretable special case.}
For $Y_t=X_0$, the optimal target is $m_t(X_t)=\mathbb E[X_0\mid X_t,t]$. Near the low-noise endpoint, $m_t(X_t)\approx X_0$, so target identifiability approximately becomes clean-tuple identifiability. Suppose each clean sample lies on a smooth $d$-dimensional manifold $\mathcal M\subset\mathbb R^D$. A tuple lies on $\mathcal M^K$, of dimension $Kd$, while the fused representation has dimension $D$. Writing $t=0$ for the clean endpoint in this limiting argument, $m_0(X_0)=X_0$. Suppose smooth exact recovery holds on a local patch of the clean manifold, so that $g_0\circ F_K$ equals the identity there. For any tangent vector $v\in T_{X_0}\mathcal M^K$, differentiating this identity gives
\begin{equation}
\mathrm d g_0\big|_{F_K(X_0)}\!\left(\mathrm d F_K\big|_{X_0}v\right)=v.
\label{eq:tangent_recovery}
\end{equation}
Here $\mathrm d$ denotes the differential of the indicated map. If fusion sent a nonzero tangent vector to zero, the left side would vanish while the right side would not. Its tangent restriction must therefore preserve all $Kd$ independent directions. Since its output lies in $\mathbb R^D$, the rank inequality is
\begin{equation}
Kd=\operatorname{rank}(J_{\mathcal M}m_0)
\le\operatorname{rank}(J_{\mathcal M}F_K)\le D.
\label{eq:clean_capacity}
\end{equation}
Here $J_{\mathcal M}$ denotes restriction to $T_{X_0}\mathcal M^K$. Hence $D\ge Kd$ is a necessary capacity condition for smooth local recovery of the complete clean tuple. When $D<Kd$, at least one intrinsic direction must be lost. When $D\ge Kd$, preserving these directions becomes dimensionally possible. This distinction motivates the controlled toy experiment, which varies $d$, $D$, and $K$. Actual recovery additionally depends on the local separation developed by the fusion map. For \algacronym{}-AE, apply the same argument to a smooth deterministic tuple-to-latent map $B_K:\mathcal M^K\to\mathbb R^{D_{\mathrm{AE}}}$. An ideal smooth decoder satisfying $\mathcal D_K\circ B_K=\mathrm{id}$ would require $D_{\mathrm{AE}}\ge Kd$. Here $D_{\mathrm{AE}}$ denotes the shared latent dimension. This applies the same capacity principle to joint reconstruction, consistent with the placement of fusion and unfusion around the autoencoder.

Stable local unfusion also calls for distinguishability of nearby clean tuples, for example
\begin{equation}
\|F_K(X)-F_K(X')\|\ge c\|X-X'\|,\qquad c>0,
\label{eq:local_distinguishability}
\end{equation}
within a local neighborhood of $\mathcal M^K$. The bound gives an inverse sensitivity of at most $1/c$ on the local image, so that separated clean tuples remain separated in fused space. To see why full tangent rank yields a local inverse, choose local coordinates $u\in\mathbb R^{Kd}$ on $\mathcal M^K$. A rank-$Kd$ derivative contains a nonsingular $Kd\times Kd$ minor. Selecting those output coordinates gives a map with an invertible derivative, so the inverse function theorem supplies a local inverse for the selected map and hence for $F_K$ on its image. Shrinking to a neighborhood with a bounded inverse derivative gives a finite local inverse Lipschitz constant, equivalent to a lower bound of the form above. The constant is local and can deteriorate as the derivative approaches rank deficiency. Slot-specific mappings $F_K(X)=\sum_{i=1}^K\psi_{K,i}(x^i)$ allow each sample to occupy different representation directions. Prediction training favors arrangements from which the slot-wise targets can be recovered. The capacity condition and this local separation condition play complementary roles. The toy, response, and weight-orthogonality experiments examine their learned counterparts.

\paragraph{Noise, target compatibility, and TFS.}
At noisy timesteps, consider $X_t=a_tX_0+\sigma_t\epsilon$ with $\sigma_t>0$ and $\epsilon\sim\mathcal N(0,I_{KD})$ independent of $X_0$. Conditional on any clean tuple, the Gaussian perturbation has positive density throughout $\mathbb R^{KD}$, so its mixture over clean tuples also has full support there. Hence the noisy input distribution occupies $KD$ ambient dimensions even though the clean tuple lies on the $Kd$-dimensional manifold $\mathcal M^K$. For $K>1$, the continuous map $F_K:\mathbb R^{KD}\rightarrow\mathbb R^D$ cannot be globally injective. To see this, an injective map could be composed with the coordinate inclusion of $\mathbb R^D$ into $\mathbb R^{KD}$. Invariance of domain would require the resulting image to be open in $\mathbb R^{KD}$, contradicting its containment in a lower-dimensional coordinate subspace. The relevant objects are its fibers
\begin{equation}
F_K^{-1}(z)=\{X\in\mathbb R^{KD}:F_K(X)=z\}.
\end{equation}
Low fusion ambiguity requires compatible optimal targets within these fibers. When velocity is the target, this recovers the original fiber-wise vector-field consistency interpretation, in which the conditional mean slot-wise velocity should have small variation within a fiber. The relevant variation is measured across tuples identified by fusion, which may be distant in input space. This motivates comparing slot-wise fields for near-fiber pairs against pairs separated in fused space. Together with alignment to a single-sample reference, these comparisons examine whether sharing a representation is accompanied by compatible prediction behavior.

Noise and SNR affect $\Delta_t(K)$ through both the conditional distribution and the target parameterization. For example, under $X_t=a_tX_0+\sigma_t\epsilon$ with $\sigma_t>0$, rearranging the noising equation gives $\epsilon=(X_t-a_tX_0)/\sigma_t$. Conditioning on $X_t$ leaves $X_t$, $a_t$, and $\sigma_t$ fixed, so linearity of conditional expectation gives
\begin{equation}
m_t^\epsilon(X_t)=\frac{X_t-a_t\mathbb E[X_0\mid X_t,t]}{\sigma_t}.
\label{eq:noise_target_information}
\end{equation}
In the pure-noise limit $a_t=0$, $\sigma_t=1$, this target equals $X_t$. This example clarifies why scheduling depends on the prediction target as well as SNR. Even in a high-noise state, direct noise prediction can require detailed state information. In comparison, a clean-state target depends on the posterior mean of $X_0$. TFS must therefore balance the information demanded by the chosen target with the effect of prediction error on the subsequent update.

TFS uses this distinction to allocate shared computation along the trajectory. For a velocity predictor, compare $\widehat X_{t'}=X_t+h_t\widehat v_t(X_t)$ and $X^*_{t'}=X_t+h_tv_t^*(X_t)$ from the same state, with signed interval $h_t=t'-t$. Subtraction cancels $X_t$, and homogeneity of the norm gives
\begin{equation}
\|\widehat X_{t'}-X^*_{t'}\|
=|h_t|\,\|\widehat v_t(X_t)-v_t^*(X_t)\|,
\label{eq:euler_prediction_error}
\end{equation}
where $X^*_{t'}$ is the Euler update using the reference velocity $v_t^*$ from the same input state. The step size directly scales the contribution of velocity error to the next state. This supplies the rationale for assigning larger fusion factors to smaller Flow Matching intervals. For DDPM/DDIM, coarse-to-fine generation motivates retaining more per-sample capacity near final refinement. The trajectory-dependent alignment in Fig.~\ref{fig:analysis} and the schedule ablations evaluate this allocation principle in the learned models. When comparing prediction risks across fusion factors, per-slot normalization such as $\Delta_t(K)/K$ accounts for the changing tuple size.

\paragraph{Conditional grouping and slot separation.}
Fiber consistency compares different complete tuples mapped to a shared representation. Conditional grouping determines how the individual slots within each tuple can be separated by unfusion. Diverse class conditions supply additional cues for assigning predicted features to the appropriate slot. The benefit of mixed-condition grouping therefore complements compatibility across tuples. One concerns slot separation, while the other concerns shared prediction behavior. Same-condition samples retain distinct noisy states and targets, providing alternative cues for unfusion. Our class-conditional grouping experiment examines how the availability of these cues affects generation quality.

\paragraph{Scope of the analysis.}
The identities above describe fixed-timestep population risk under squared loss, and the clean-capacity argument assumes smooth maps on the clean manifold. The empirical analyses use learned reference fields and finite perturbations to examine the associated prediction geometry. End-to-end generation quality and the cumulative effects of fusion are assessed by the sampling metrics and TFS ablations.

\FloatBarrier
\section{Empirical Geometric Analysis}
\paragraph{Toy experiment}

To empirically validate the proposed geometric analysis, we construct a controlled setting in which the intrinsic dimension $d$ and ambient dimension $D$ are explicitly separated. Latent samples $\hat{x}\in\mathbb{R}^d$ are drawn from a low-dimensional distribution and embedded into the ambient space $\mathbb{R}^D$ through a fixed column-orthogonal projection $P\in\mathbb{R}^{D\times d}$ satisfying $P^\top P=I$, yielding observations $x=P\hat{x}$. The projection is unknown to the model, so diffusion models are trained entirely in the ambient space using lightweight MLP denoisers. To inspect the intrinsic structure, generated samples are projected back to the latent coordinates through $P^\top$. Within this setting, we train denoisers both with and without fusion and analyze the resulting fused dynamics under different fusion factors.

\begin{figure}[!htbp]
    \centering
    \includegraphics[width=1\linewidth]{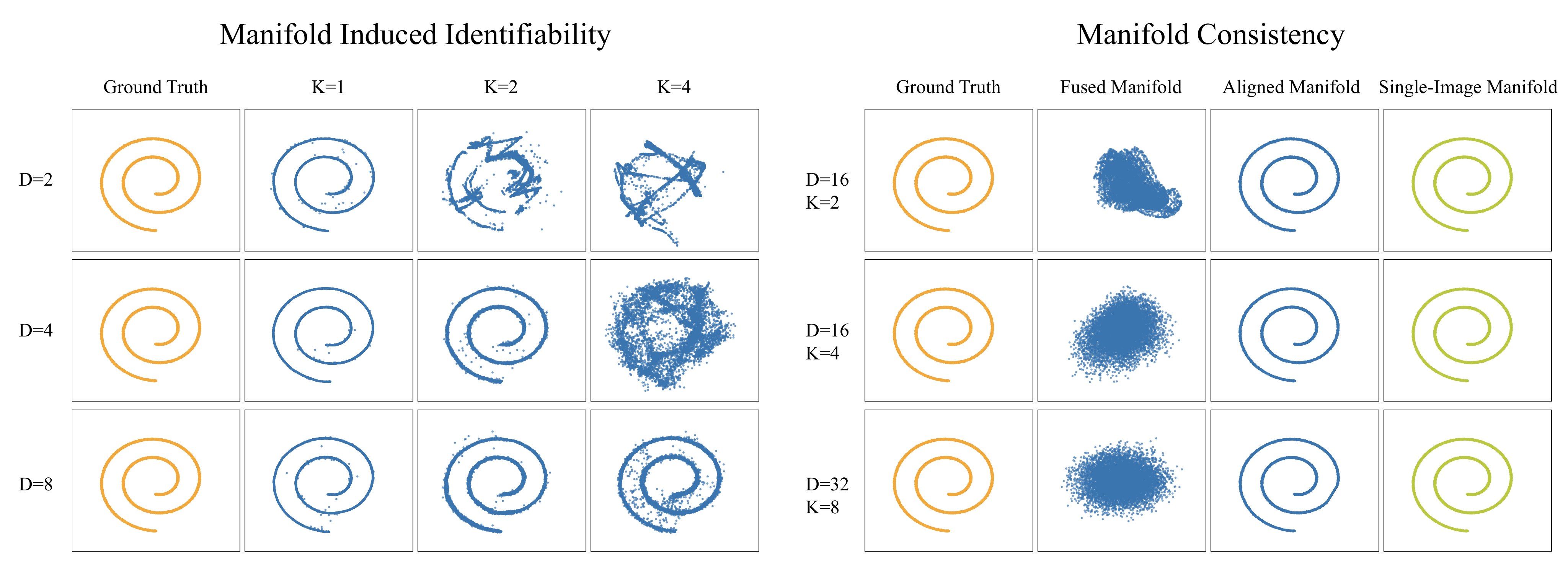}
    \caption{
    Clean-capacity and representation-alignment diagnostics under multi-sample fusion.
    \textit{Left.} Swiss-roll samples as ambient dimension $D$ and fusion factor $K$ vary.
    \textit{Right.} An affine transformation aligns the visualized fused representations with the single-image reference in this toy setting. The RMSEs are $4.83\times10^{-4}$ for $(D{=}16,K{=}2)$, $7.03\times10^{-4}$ for $(D{=}16,K{=}4)$, and $7.8\times10^{-5}$ for $(D{=}32,K{=}8)$.
    }
    \label{fig:toy}
\end{figure}

To examine clean-manifold capacity under fusion, we vary the relationship between the ambient dimension $D$ and the fused intrinsic dimension $Kd$. Consistent with the necessary capacity condition derived in Sec.~\ref{subsec:geometric_analysis}, recovery fails when the ambient space cannot represent the fused product manifold, while stable separation is observed for $D\ge Kd$ in the evaluated configurations (Fig.~\ref{fig:toy}). This transition empirically supports the role of representation capacity in clean-sample recovery.

To examine representation alignment, we analyze the geometry induced by the fusion map in the ambient denoising space. We visualize fused representations after projection to the intrinsic coordinates through $P^\top$. Although the fused geometry does not initially coincide with the single-sample manifold, an affine alignment in the ambient space substantially reduces the discrepancy. In recoverable regimes, the aligned fused geometry closely matches the single-sample manifold structure, supporting preservation of the reference structure in the projected features.

\FloatBarrier
\paragraph{Local Smoothness and Empirical Lipschitz Analysis}

We further examine the local smoothness of the denoising dynamics through empirical local response ratios. For a denoiser $f_t$, we measure
\begin{equation}
\frac{\|f_t(x+\delta)-f_t(x)\|_2}{\|\delta\|_2},
\end{equation}
where $\delta$ denotes a small local perturbation around the current noisy sample. This ratio measures how sensitively the denoising output changes in the sampled perturbation direction.

\begin{figure}[!htbp]
    \centering
    \includegraphics[width=1\linewidth]{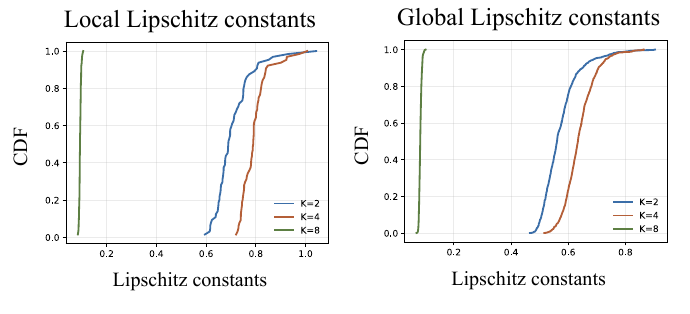}
    \caption{
    Empirical local response ratios under multi-sample fusion.
    We measure local Lipschitz ratios of the denoising dynamics under both near-neighbor and fusion-induced perturbations across fusion factors.
    The estimated ratios for $K{=}2$ and $K{=}4$ remain clearly positive while staying well controlled, showing nonzero output responses in the sampled directions.
    For $K{=}8$, the ratio remains positive but becomes noticeably smaller, indicating attenuated responses under more aggressive fusion in this diagnostic.
    }
    \label{fig:lipschitz}
\end{figure}

We observe that the empirical ratios for $K{=}2$ and $K{=}4$ remain clearly above zero while staying well controlled, indicating that the denoiser remains responsive to the sampled input variations (Fig.~\ref{fig:lipschitz}). For $K{=}8$, the ratio is still positive, but noticeably smaller than those of $K{=}2$ and $K{=}4$. This suggests that local prediction differences become more attenuated under aggressive fusion. These results support the TFS design, which reduces fusion during sensitive refinement stages.

\FloatBarrier
\paragraph{Interpolation Behavior under \algname{}}
\label{subsec:interp}

\begin{figure}[!htbp]
    \centering
    \includegraphics[width=0.5\linewidth]{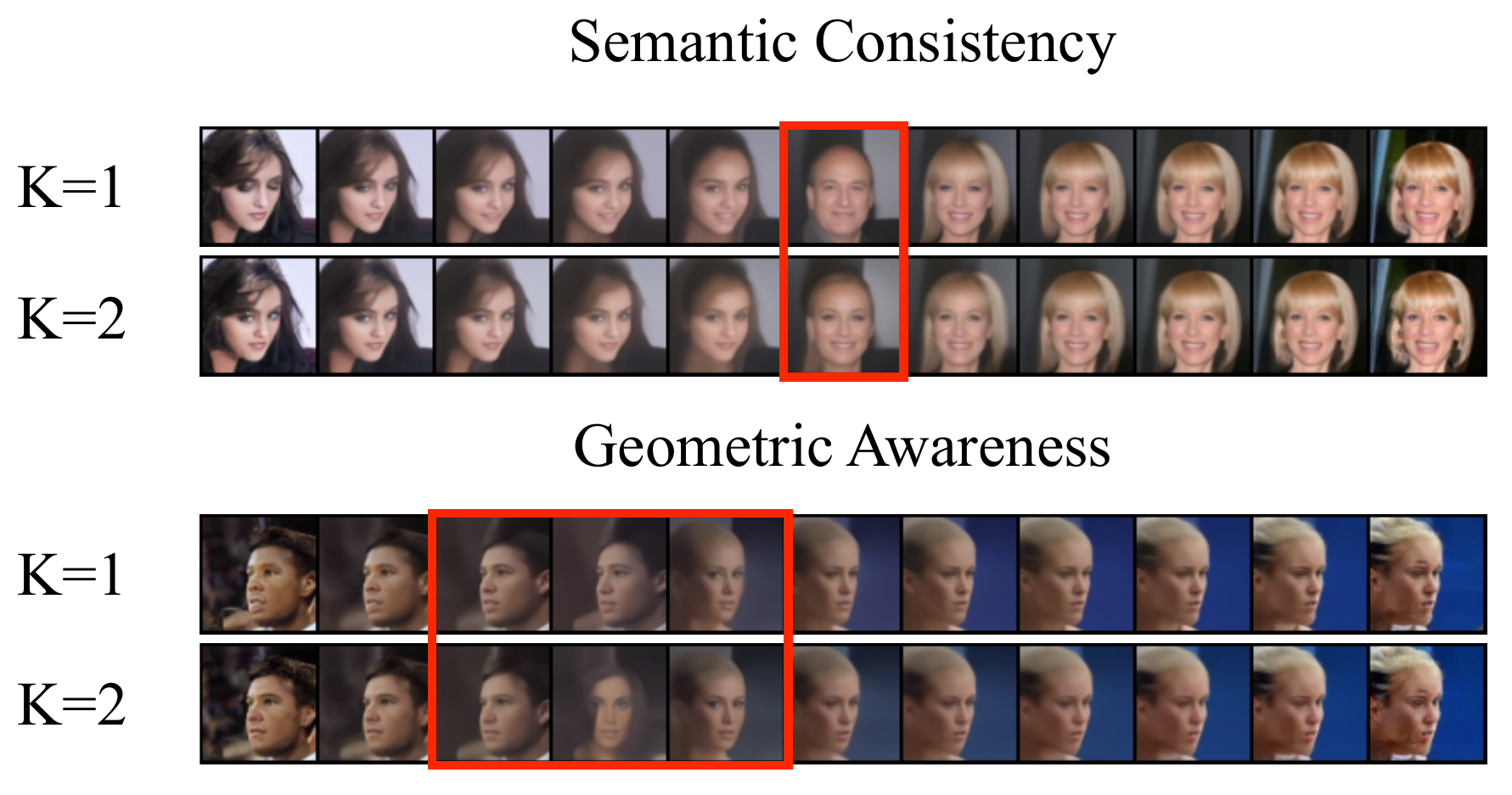}
    \caption{Interpolation behavior under ManiFusion on CelebA. The top panel shows semantic interpolation between two identities by interpolating noise and running the reverse process. The bottom panel shows geometric interpolation between pose/appearance factors. Compared with the baseline (K = 1), ManiFusion (K = 2) yields smoother transitions and fewer abrupt mode switches. Red boxes highlight representative transition regions.
}
    \label{fig:interpolation}
\end{figure}

We analyze linear interpolation in the noise space. We randomly sample two CelebA images $x_0^a, x_0^b$ and map them to the terminal diffusion time $t{=}T$ via the forward process, yielding $x_T^a$ and $x_T^b$. We then construct interpolated noises
\begin{equation}
\tilde x_T(\lambda)=(1-\lambda)x_T^a+\lambda x_T^b,\quad \lambda\in[0,1],
\end{equation}
and run the reverse process starting from $\tilde x_T(\lambda)$.
We compare a standard baseline and \algname{} ($K{=}2$). We visualize the resulting samples $\hat x_0(\lambda)$ along a grid of $\lambda$ values.
In Fig.~\ref{fig:interpolation}, baseline often exhibits mode switching and non-smooth transitions. In contrast, the \algname{} produces noticeably smoother morphing between the two identities, showing smoother interpolation behavior. These examples support coherent changes in generated appearance as the input noise varies.

\FloatBarrier
\paragraph{Manifold Coverage under \algname}

\begin{figure}[!htbp]
    \centering
    \includegraphics[width=0.8\linewidth]{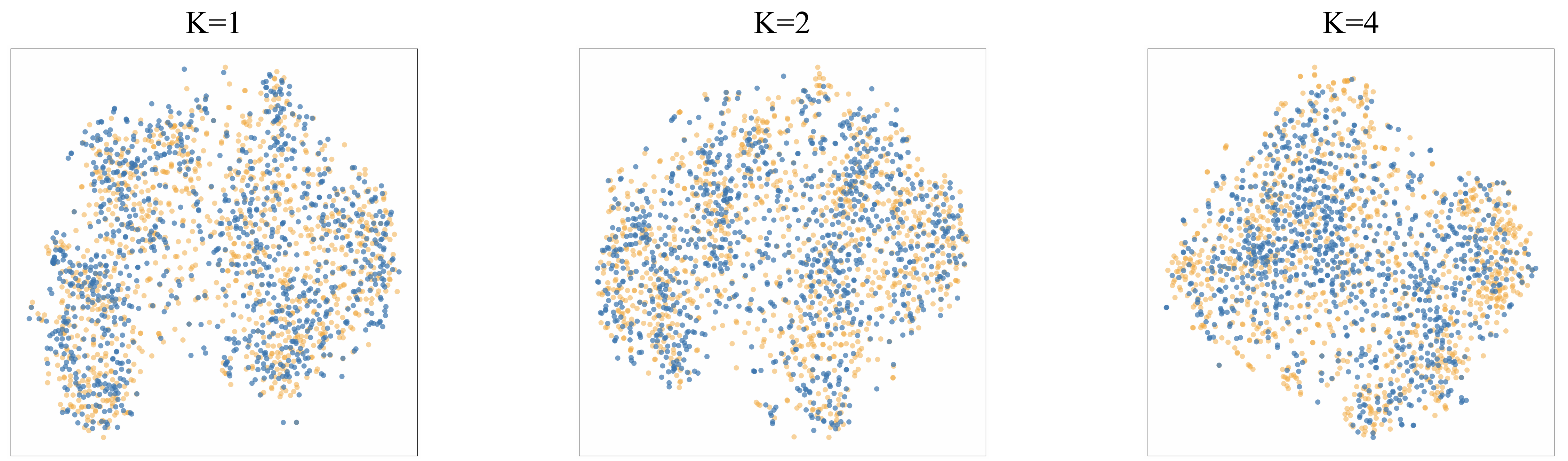}
    \caption{t-SNE visualization of feature distributions on CIFAR-10. The plots visualize the feature embeddings of real data (orange) and generated samples (blue) across different fusion factors.
  The panels show standard diffusion ($K=1$) and ManiFusion with $K=2$ and $K=4$ from left to right.}
    \label{fig:tsne}
\end{figure}

While \algname\ slightly reduces precision, recall remains comparable to the baseline model. This suggests that the generative model continues to cover similar regions of the evaluation feature space despite signal fusion. To examine this hypothesis, we visualize real and generated samples in feature space using t-SNE. The visualization shows that samples from \algname\ occupy a similar region of the displayed feature space as those from the baseline model, indicating comparable coverage(Fig.~\ref{fig:tsne}). Together, recall and these visualizations support comparable sample diversity in the evaluated setting.

\clearpage
\section{Extensions of \algname{}}
\paragraph{\textbf{Large-Fusion Extension and Trajectory-Fusion Scheduling}}
We extend the fusion factor to larger values and study TFS across all experiments in Sec~\ref{sec:Method}. Specifically, we consider fusion factors up to $K=8$ and evaluate schedules that combine $\{8,4,2,1\}$ with different ratios across denoising timesteps. As shown in Table~\ref{tab:appen_tfs}, Table~\ref{tab:deepcache_appen} and Table~\ref{tab:imagenet_fb_appen}, naive fusion with large $K$ significantly degrades generation quality despite providing large throughput gains. \textit{However, for \algacronym{}-DiT, compared with increasing patch size for efficiency as in DiT-S/4, \algname{} (-TFS, K=4) achieves similar throughput gains while preserving substantially better sample quality.} Moreover, \algname{} reduces the fusion factor over the denoising process and enables the use of larger fusion factors without severe quality degradation. Notably, \textit{using $K=8$ with a uniform schedule $(25\%,25\%,25\%,25\%)$ achieves both higher throughput and better generation quality compared to schedules that rely only on $\{4,2,1\}$}. This result highlights that large fusion factors early in denoising can further improve the efficiency–fidelity trade-off.

\begin{table*}[!htbp]
    \centering
    \caption{Extended comparison of Vanilla, Ours (-TFS), and \textbf{Ours} across DDPM, DDIM, FM, and DiT on CIFAR-10 and CelebA. Ours (-TFS) uses a fixed fusion factor \(K\), while \textbf{Ours} dynamically allocates fusion factors across denoising timesteps. For \textbf{Ours}, each tuple specifies the timestep allocation ratio for \(K=\{8,4,2,1\}\). Larger fusion factors and additional TFS schedules are evaluated only for DDPM. Best results are in \textbf{bold}, and second-best results are \underline{underlined}.}
    \vspace{1mm}
    \label{tab:appen_tfs}

    \setlength{\tabcolsep}{3.5pt}
    \renewcommand{\arraystretch}{1.08}

    \begin{adjustbox}{max width=\linewidth}
    \begin{tabular}{l|l|cc|c|ccccc}
        \toprule
        \multicolumn{10}{c}{\textbf{CIFAR-10}} \\
        \midrule
        Method & Variant & Throughput$\uparrow$ & MAC$\downarrow$ & Parameter & FID$\downarrow$ & sFID$\downarrow$ & IS$\uparrow$ & Precision$\uparrow$ & Recall$\uparrow$ \\
        \midrule

        \multirow{7}{*}{DDPM}
          & Vanilla & 1.71 & 6.07T & 35.7M & \textbf{3.41} & 4.40 & \underline{9.20} & \textbf{0.68} & \underline{0.58} \\
          & Ours (-TFS, \(K=2\)) & 3.17 ($\times1.85$) & 3.07T (50.6\%) & +63.1K (0.18\%) & 5.93 & 4.58 & 8.67 & 0.63 & \underline{0.58} \\
          & Ours (-TFS, \(K=4\)) & \underline{6.31} ($\times3.69$) & \underline{1.55T} (25.5\%) & +102.0K (0.29\%) & 18.27 & 6.54 & 7.41 & 0.56 & 0.55 \\
          & Ours (-TFS, \(K=8\)) & \textbf{10.36} ($\times6.06$) & \textbf{0.78T} (12.9\%) & +179.8K (0.50\%) & 43.53 & 13.95 & 5.47 & 0.62 & 0.45 \\
          & \textbf{Ours} (0\%, 50\%, 25\%, 25\%) & 3.31 ($\times1.94$) & 3.06T (50.4\%) & +102.0K (0.29\%) & 3.88 & \textbf{4.27} & \textbf{9.29} & \underline{0.66} & \textbf{0.60} \\
          & \textbf{Ours} (25\%, 25\%, 25\%, 25\%) & 3.82 ($\times2.23$) & 2.87T (47.2\%) & +179.8K (0.50\%) & \underline{3.87} & \underline{4.28} & \textbf{9.29} & \underline{0.66} & \textbf{0.60} \\
          & \textbf{Ours} (50\%, 25\%, 12.5\%, 12.5\%) & 5.00 ($\times2.92$) & 1.92T (31.7\%) & +179.8K (0.50\%) & 8.53 & 5.25 & 8.74 & 0.61 & \textbf{0.60} \\
        \midrule

        \multirow{4}{*}{DDIM (T=50)}
          & Vanilla & 35.47 & 303.44G & 35.7M & \textbf{7.99} & \underline{6.66} & \textbf{8.71} & \textbf{0.64} & 0.55 \\
          & Ours (-TFS, \(K=2\)) & 65.39 ($\times1.84$) & 153.44G (50.6\%) & +63.1K (0.18\%) & 13.34 & 7.16 & 8.11 & 0.59 & \underline{0.56} \\
          & Ours (-TFS, \(K=4\)) & \textbf{120.95} ($\times3.41$) & \textbf{77.31G} (25.5\%) & +102.0K (0.29\%) & 26.92 & 7.95 & 7.16 & 0.53 & 0.54 \\
          & \textbf{Ours} (0\%, 50\%, 25\%, 25\%) & \underline{66.80} ($\times1.88$) & \underline{152.87G} (50.4\%) & +102.0K (0.29\%) & \underline{10.35} & \textbf{6.29} & \underline{8.58} & \underline{0.60} & \textbf{0.57} \\
        \midrule

        \multirow{4}{*}{FM}
          & Vanilla & 9.60 & 124.55G & 55.7M & \textbf{2.74} & \textbf{3.84} & \textbf{9.47} & \textbf{0.65} & \textbf{0.63} \\
          & Ours (-TFS, \(K=2\)) & 18.22 ($\times1.90$) & 62.89G (50.5\%) & +185.9K (0.33\%) & 6.72 & 4.50 & \underline{8.76} & \underline{0.62} & \underline{0.60} \\
          & Ours (-TFS, \(K=4\)) & \textbf{32.98} ($\times3.44$) & \textbf{31.69G} (25.4\%) & +323.4K (0.58\%) & 21.11 & 7.10 & 7.35 & 0.58 & 0.53 \\
          & \textbf{Ours} (0\%, 50\%, 25\%, 25\%) & \underline{18.29} ($\times1.91$) & \underline{62.70G} (50.3\%) & +323.4K (0.58\%) & \underline{4.10} & \underline{3.95} & 8.72 & 0.61 & 0.59 \\
        \midrule

        \multirow{4}{*}{DiT}
          & Vanilla & 8.25 & 217.00G & 32.48M & \textbf{10.58} & \underline{7.37} & \textbf{10.17} & \textbf{0.75} & \textbf{0.40} \\
          & Ours (-TFS, \(K=2\)) & 15.86 ($\times1.92$) & 114.25G (52.6\%) & +1.04M (3.02\%) & 11.13 & \textbf{6.45} & 9.50 & 0.72 & \textbf{0.40} \\
          & Ours (-TFS, \(K=4\)) & \textbf{29.88} ($\times3.62$) & \textbf{60.00G} (27.6\%) & +1.63M (5.02\%) & 18.59 & 8.68 & 8.74 & 0.68 & 0.34 \\
          & \textbf{Ours} (0\%, 50\%, 25\%, 25\%) & \underline{15.95} ($\times1.93$) & \underline{113.84G} (52.5\%) & +1.63M (5.02\%) & \underline{10.97} & 7.71 & \underline{9.60} & \underline{0.73} & \underline{0.37} \\

        \midrule
        \multicolumn{10}{c}{\textbf{CelebA}} \\
        \midrule

        Method & Variant & Throughput$\uparrow$ & MAC$\downarrow$ & Parameter & FID$\downarrow$ & sFID$\downarrow$ & IS$\uparrow$ & Precision$\uparrow$ & Recall$\uparrow$ \\
        \midrule

        \multirow{7}{*}{DDPM}
          & Vanilla & 0.71 & 15.56T & 113.7M & \textbf{3.51} & \underline{6.27} & \underline{2.77} & \textbf{0.76} & 0.50 \\
          & Ours (-TFS, \(K=2\)) & 1.28 ($\times1.80$) & 7.92T (50.9\%) & +63.1K (0.06\%) & 4.73 & 6.98 & 2.70 & \underline{0.75} & 0.46 \\
          & Ours (-TFS, \(K=4\)) & \underline{2.34} ($\times3.30$) & \underline{4.01T} (25.8\%) & +102.0K (0.09\%) & 9.15 & 10.34 & 2.64 & 0.70 & 0.40 \\
          & Ours (-TFS, \(K=8\)) & \textbf{3.84} ($\times5.41$) & \textbf{2.04T} (13.1\%) & +179.8K (0.16\%) & 34.60 & 30.21 & 2.26 & 0.56 & 0.11 \\
          & \textbf{Ours} (0\%, 50\%, 25\%, 25\%) & 1.33 ($\times1.87$) & 7.87T (50.6\%) & +102.0K (0.09\%) & \underline{3.81} & \textbf{6.00} & \underline{2.77} & 0.74 & \underline{0.51} \\
          & \textbf{Ours} (25\%, 25\%, 25\%, 25\%) & 1.40 ($\times1.97$) & 7.38T (47.4\%) & +179.8K (0.16\%) & 3.97 & 7.02 & \textbf{2.83} & 0.73 & \textbf{0.52} \\
          & \textbf{Ours} (50\%, 25\%, 12.5\%, 12.5\%) & 1.98 ($\times2.79$) & 4.96T (31.9\%) & +179.8K (0.16\%) & 7.72 & 8.51 & 2.76 & 0.67 & 0.49 \\
        \midrule

        \multirow{4}{*}{DDIM (T=50)}
          & Vanilla & 14.56 & 778.00G & 113.7M & \textbf{5.58} & \underline{8.37} & \underline{2.81} & \textbf{0.74} & \underline{0.46} \\
          & Ours (-TFS, \(K=2\)) & 26.05 ($\times1.79$) & 395.88G (50.1\%) & +63.1K (0.06\%) & 7.43 & 9.07 & 2.75 & \underline{0.71} & 0.42 \\
          & Ours (-TFS, \(K=4\)) & \textbf{47.67} ($\times3.27$) & \textbf{200.32G} (25.8\%) & +102.0K (0.09\%) & 12.62 & 12.26 & 2.67 & 0.65 & 0.36 \\
          & \textbf{Ours} (0\%, 50\%, 25\%, 25\%) & \underline{26.83} ($\times1.84$) & \underline{393.63G} (50.6\%) & +102.0K (0.09\%) & \underline{6.52} & \textbf{7.82} & \textbf{2.84} & 0.68 & \textbf{0.48} \\
        \midrule

        \multirow{4}{*}{FM}
          & Vanilla & 2.12 & 589.55G & 169.1M & \textbf{2.80} & \underline{5.35} & \textbf{3.23} & \underline{0.67} & \textbf{0.60} \\
          & Ours (-TFS, \(K=2\)) & 4.06 ($\times1.92$) & 297.24G (50.4\%) & +185.9K (0.11\%) & 3.67 & 5.67 & 3.17 & \textbf{0.68} & 0.56 \\
          & Ours (-TFS, \(K=4\)) & \textbf{7.65} ($\times3.61$) & \textbf{149.59G} (25.4\%) & +323.4K (0.19\%) & 7.15 & 7.24 & 3.00 & 0.62 & 0.52 \\
          & \textbf{Ours} (0\%, 50\%, 25\%, 25\%) & \underline{4.08} ($\times1.92$) & \underline{296.49G} (50.3\%) & +323.4K (0.19\%) & \underline{2.98} & \textbf{5.20} & \underline{3.22} & 0.66 & \underline{0.59} \\
        \bottomrule
    \end{tabular}
    \end{adjustbox}
\end{table*}

\clearpage
\paragraph{Combination with DeepCache}
Table~\ref{tab:deepcache_appen} reports the combined throughput and quality of fusion and DeepCache, with and without TFS.

\begin{table*}[!htbp]
\centering
\caption{Combination of \algname{} (\algacronym{}-UNet) w/w.o TFS with DeepCache on CIFAR-10 and CelebA. MAC reduction ratios and throughput speedups are shown in parentheses.
\textsuperscript{*}DDIM steps are reduced to match the same throughput setting.}
\vspace{1mm}
\label{tab:deepcache_appen}
\begin{adjustbox}{max width=\linewidth}
\begin{tabular}{l|l|ll|ccccc}
\toprule
\multicolumn{9}{c}{\textbf{CIFAR-10}} \\
\midrule
\multicolumn{1}{c}{Setting} &
\multicolumn{1}{c}{Variant} &
\multicolumn{1}{c}{Throughput$\uparrow$} &
\multicolumn{1}{c}{MAC$\downarrow$ (G)} &
\multicolumn{1}{c}{FID$\downarrow$} &
\multicolumn{1}{c}{sFID$\downarrow$} &
\multicolumn{1}{c}{IS$\uparrow$} &
\multicolumn{1}{c}{Prec$\uparrow$} &
\multicolumn{1}{c}{Rec$\uparrow$} \\
\midrule
\multirow[c]{2}{*}{DDIM} & Vanilla & 17.70 & 606.88 & 6.30 & 5.65 & 8.88 & 0.65 & 0.56 \\
 & +DeepCache & 28.05 ($\times$1.59) & 390.18 (64.3\%) & 6.90 & 6.21 & 8.89 & 0.64 & 0.56 \\
 &  \; +Step-Reduction & 49.02\textsuperscript{*} ($\times$2.77) & 207.88 (34.3\%) & 10.12 & 7.40 & 8.56 & 0.63 & 0.53 \\
\midrule
\multirow[c]{2}{*}{Ours (-TFS, $K=2$)} & Vanilla & 33.48 ($\times$1.89) & 306.88 (50.6\%) & 11.21 & 6.12 & 8.30 & 0.59 & 0.57 \\
 & +DeepCache & 51.19 ($\times$2.89) & 198.48 (32.7\%) & 11.24 & 6.48 & 8.29 & 0.59 & 0.56 \\
\midrule
\multirow[c]{2}{*}{Ours (-TFS, $K=4$)} & Vanilla & 56.80 ($\times$3.21) & 154.63 (25.5\%) & 23.72 & 7.22 & 7.31 & 0.54 & 0.55 \\
 & +DeepCache & 91.88 ($\times$5.19) & 100.43 (16.6\%) & 23.87 & 8.11 & 7.37 & 0.52 & 0.54 \\
\midrule
\multirow[c]{2}{*}{\textbf{Ours}} & Vanilla & 32.90 ($\times$1.86) & 307.76 (50.7\%) & 8.65 & 5.46 & 8.68 & 0.61 & 0.59 \\
 & +DeepCache & 51.98\textsuperscript{*} ($\times$2.94) & 197.38 (32.5\%) & 9.42 & 6.37 & 8.62 & 0.61 & 0.57 \\
\midrule
\multicolumn{9}{c}{\textbf{CelebA}} \\
\midrule
\multicolumn{1}{c}{Setting} &
\multicolumn{1}{c}{Variant} &
\multicolumn{1}{c}{Throughput$\uparrow$} &
\multicolumn{1}{c}{MAC$\downarrow$ (G)} &
\multicolumn{1}{c}{FID$\downarrow$} &
\multicolumn{1}{c}{sFID$\downarrow$} &
\multicolumn{1}{c}{IS$\uparrow$} &
\multicolumn{1}{c}{Prec$\uparrow$} &
\multicolumn{1}{c}{Rec$\uparrow$} \\
\midrule
\multirow[c]{2}{*}{DDIM} & Vanilla & 7.20 & 1556.01 & 4.17 & 6.70 & 2.82 & 0.74 & 0.50 \\
 & +DeepCache & 9.81 ($\times$1.36) & 1075.81 (69.1\%) & 6.07 & 9.10 & 2.79 & 0.72 & 0.45 \\
 &  \; +Step-Reduction & 16.83\textsuperscript{*} ($\times$2.34) &  572.58 (36.8\%) & 8.22 & 11.38 & 2.73 & 0.70 & 0.40 \\
\midrule
\multirow[c]{2}{*}{Ours (-TFS, $K=2$)} & Vanilla & 13.25 ($\times$1.84) & 791.76 (50.9\%) & 6.17 & 7.71 & 2.76 & 0.71 & 0.45 \\
 & +DeepCache & 16.96 ($\times$2.36) & 551.66 (35.5\%) & 8.43 & 10.50 & 2.71 & 0.69 & 0.40 \\
\midrule
\multirow[c]{2}{*}{Ours (-TFS, $K=4$)} & Vanilla & 23.81 ($\times$3.31) & 400.63 (25.8\%) & 12.07 & 11.93 & 2.69 & 0.65 & 0.38 \\
 & +DeepCache & 30.41 ($\times$4.23) & 280.63 (18.0\%) & 14.29 & 14.03 & 2.62 & 0.63 & 0.32 \\
\midrule
\multirow[c]{2}{*}{\textbf{Ours}} & Vanilla & 13.41 ($\times$1.86) & 787.26 (50.6\%) & 5.54 & 6.91 & 2.85 & 0.69 & 0.51 \\
 & +DeepCache & 17.65\textsuperscript{*} ($\times$2.45) & 547.18 (35.2\%) & 7.07 & 8.69 & 2.79 & 0.68 & 0.46 \\
\bottomrule
\end{tabular}
\end{adjustbox}
\end{table*}

\FloatBarrier
\paragraph{Scaling Fusion on ImageNet}
Table~\ref{tab:imagenet_fb_appen} compares fixed fusion factors and TFS on ImageNet using DiT-S/2, alongside the DiT-S/4 baseline.

\begin{table}[!htbp]
\centering
\caption{Extension of Ours(-TFS) to larger fusion factors and \textbf{Ours} on ImageNet using a DiT-S/2 backbone. Numbers in parentheses indicate throughput speedups and MAC reduction ratios relative to the baseline. For \textbf{Ours}, each tuple (e.g., $(0\%,50\%,25\%,25\%)$) specifies the fraction of denoising timesteps allocated to fusion factors $K=\{8,4,2,1\}$.}
\vspace{1mm}
\label{tab:imagenet_fb_appen}
\begin{adjustbox}{max width=\linewidth}
\begin{tabular}{l | l l | c | c c c c c}
\toprule
\multicolumn{9}{c}{\textbf{ImageNet}} \\
\midrule
Method & Throughput$\uparrow$ & MAC$\downarrow$ (G) & Parameter & FID$\downarrow$ & sFID$\downarrow$ & IS$\uparrow$ & Prec$\uparrow$ & Rec$\uparrow$ \\
\midrule
DiT-S/2 (baseline) & 3.04 & 1067.50 & 32.9M & 71.85 & 12.29 & 18.65 & 0.35 & 0.55 \\
\midrule
Ours (-TFS,K=2) & 5.95 ($\times$1.96) & 562.50 (52.69\%) & +1.33M (4.04\%) & 78.22 & 15.38 & 16.83 & 0.34 & 0.53 \\
Ours (-TFS,K=4) & 11.61 ($\times$3.82) & 286.88 (26.87\%) & +1.94M (5.90\%) & 93.64 & 20.59 & 14.30 & 0.26 & 0.33 \\
Ours (-TFS,K=8) & 21.42 ($\times$7.05) & 153.75 (14.40\%) & +2.56M (7.78\%) & 136.06 & 46.86 & 8.14 & 0.16 & 0.13 \\
\midrule
\textbf{Ours} (0\%, 50\%, 25\%, 25\%) & 5.98 ($\times$1.97) & 551.95 (51.70\%) & +1.94M (5.90\%) & 74.19 & 14.85 & 19.08 & 0.34 & 0.49 \\
\textbf{Ours} (25\%, 25\%, 25\%, 25\%) & 6.38 ($\times$2.10) & 520.03 (48.71\%) & +2.56M (7.78\%) & 79.16 & 15.62 & 16.67 & 0.34 & 0.53 \\
\textbf{Ours} (50\%, 25\%, 12.5\%, 12.5\%) & 9.39 ($\times$3.09) & 351.29 (32.90\%) & +2.56M (7.78\%) & 97.98 & 23.49 & 12.91 & 0.28 & 0.40 \\
\bottomrule
\end{tabular}
\end{adjustbox}
\end{table}

\FloatBarrier
\paragraph{\textbf{Extension to Latent Diffusion Models.}}
We examine the applicability of \algname{} to latent diffusion models by studying how fusion can be integrated into the denoising backbone. In particular, we evaluate MF-UNet, which applies fusion inside the UNet backbone, and compare it with the MF-AE design used in the main results. As shown in Table~\ref{tab:ldm_fb_unet}, \algname{} (-TFS) provides significant computational savings and throughput gains (up to $3.87\times$) but leads to noticeable degradation in generation quality as the fusion factor increases. Compared to the MF-AE design presented in the main results, MF-UNet consistently yields worse fidelity across metrics such as FID and sFID. Qualitative comparisons between MF-UNet and MF-AE are provided in Table~\ref{tab:ldm_comparison}.

\begin{table}[!htbp]
\centering
\caption{\algacronym{}-UNet in Latent Diffusion Models on CelebA-HQ.}
\vspace{1mm}
\label{tab:ldm_fb_unet}
\begin{adjustbox}{max width=\linewidth}
\begin{tabular}{l|ll|c|ccccc}
\toprule
\multicolumn{1}{c}{} & \multicolumn{8}{c}{\textbf{CelebA-HQ}} \\
\midrule
Method & Throughput & MAC(T) & Parameter & FID$\downarrow$ & sFID$\downarrow$ & IS$\uparrow$ & Prec$\uparrow$ & Rec$\uparrow$ \\
\midrule
LDM (baseline) & 0.54 & 96.38 & 362.3M & 5.10 & 10.13 & 3.28 & 0.72 & 0.49 \\
\midrule
MF-UNet ($K{=}2$) & 1.07 ($\times$1.98) & 48.67 (50.5\%) & +159.8K (0.04\%) & 11.60 & 11.87 & 3.08 & 0.62 & 0.42 \\
MF-UNet ($K{=}4$) & 2.09 ($\times$3.87) & 24.61 (25.5\%) & +319.6K (0.09\%) & 16.86 & 18.59 & 3.04 & 0.56 & 0.39 \\
\bottomrule
\end{tabular}
\end{adjustbox}
\end{table}

\FloatBarrier
\section{Analysis of Fusion Behavior}

\paragraph{\textbf{Conditional Grouping and Intra-Tuple Separability}}
We analyze how conditional grouping affects the separability of fused samples in \algacronym{}-DiT. During training, fused inputs can be constructed either by grouping samples that share the same condition or by randomly mixing conditions within a fusion group. Table~\ref{tab:appen_conditional} shows that random condition mixing consistently outperforms uni-class grouping. This observation supports the interpretation that conditional diversity provides clearer slot-specific cues and eases index-conditioned unfusion. This is an intra-tuple separation effect, whereas fiber consistency concerns compatibility between predictions for different complete tuples sharing a fused representation. Same-condition grouping remains effective in the evaluated setting because individual trajectories still have distinct noisy states and targets, although fewer condition-specific cues can make separation harder.

\begin{table}[!htbp]
\centering
\caption{Effect of conditional grouping in \algname{}.
We compare grouping samples from the same class (uni-class) versus mixing different classes within a fused group (multi-class).
Numbers in parentheses indicate throughput speedups and MAC reduction ratios relative to the baseline.}
\vspace{1mm}
\label{tab:appen_conditional}
\begin{adjustbox}{max width=\linewidth}
\begin{tabular}{l l | l l | c c c c c}
\toprule
Method & Variant & Throughput$\uparrow$ & MAC$\downarrow$ (G) & FID$\downarrow$ & sFID$\downarrow$ & IS$\uparrow$ & Prec$\uparrow$ & Rec$\uparrow$ \\
\midrule
Baseline & - & 34.21 & 108.47 & 10.58 & 7.37 & 10.17 & 0.75 & 0.40 \\
\midrule
Ours (-TFS, $K{=}2$) & uni-class & 64.87 ($\times$1.90) & 56.41 (52.0\%) & 14.89 & 8.49 & 8.87 & 0.69 & 0.37 \\
Ours (-TFS, $K{=}2$) & multi-class & 64.61 ($\times$1.89) & 56.41 (52.0\%) & 11.13 & 6.45 & 9.50 & 0.72 & 0.40 \\
\midrule
Ours (-TFS, $K{=}4$) & uni-class & 122.69 ($\times$3.58) & 29.56 (27.3\%) & 23.54 & 9.65 & 8.39 & 0.61 & 0.33 \\
Ours (-TFS, $K{=}4$) & multi-class & 122.24 ($\times$3.57) & 29.56 (27.3\%) & 18.59 & 8.68 & 8.74 & 0.68 & 0.34 \\
\bottomrule
\end{tabular}
\end{adjustbox}
\end{table}

\FloatBarrier
\section{Ablation Study}

We conduct an ablation study on the kernel size used in the fusion stage. As shown in Table~\ref{tab:kernel}, increasing the kernel size consistently degrades generation quality across FID, sFID, and IS. We attribute this behavior to stronger spatial entanglement introduced by larger kernels, which mixes information from different inputs more aggressively during fusion and makes unfusion harder. Across metrics, compact kernels reduce entanglement and preserve sample quality under fusion.

\begin{table}[!htbp]
\centering
\caption{Ablation on fusion kernel size. Larger kernels increase spatial entanglement, degrading sample quality. Smaller kernels (e.g., size 3) yield the best performance across FID, sFID, and IS.}
\vspace{1mm}
\label{tab:kernel}

\begin{tabular}{c|c c c c c}
\toprule
Kernel & FID$\downarrow$ & sFID$\downarrow$ & IS$\uparrow$ & Prec$\uparrow$ & Rec$\uparrow$ \\
\midrule
3 & \textbf{5.93} & \textbf{4.58} & \textbf{8.67} & \textbf{0.63} & 0.58 \\
5 & 6.94 & 4.74 & 8.58 & 0.61 & \textbf{0.59} \\
7 & 7.47 & 4.74 & 8.33 & 0.62 & 0.58 \\
\bottomrule
\end{tabular}

\end{table}

\clearpage
\section{Comparison with Throughput-Oriented Baselines}
\paragraph{\textbf{Comparison with a Naive Patch-Based Baseline.}}
We first compare \algname{} with a naive patch-based strategy that attempts to increase throughput by spatially compositing multiple inputs. Using the same model architectures and dataset configurations described in Section~\ref{sec:implementaion}, we design a naive multi-sample strategy where each noisy input is first downscaled to one-quarter resolution, concatenated into a \(2\times2\) grid, processed once by the diffusion model, and then split and upscaled back to full resolution. This simple approach, however, severely degrades generation quality. As illustrated in Table~\ref{tab:grid-images}, the patch-based method introduces artifacts, unnatural boundaries, and facial/background blur, while \algname{} preserves texture and color consistency.

Quantitative results in Table~\ref{tab:performance_appen} further confirm this gap. When applied to a standard DDPM, the patch-based method suffers a dramatic quality drop—on CIFAR-10, FID deteriorates from 18.27 (\algname{}) to 91.71, and IS falls from 7.41 to 4.55. On CelebA, \algname{} achieves an FID of 9.50 versus 26.07 for the grid approach, and an IS of 2.58 compared to 2.37. These findings demonstrate that our fusion–unfusion modules not only preserve and disentangle multi-sample information more effectively than naive patch compositing but also scale reliably to higher throughput settings. Together, the qualitative and quantitative evidence underscores the scalability of \algname{} and its potential to unlock further efficiency gains without compromising visual fidelity.

\begin{table}[!htbp]
\centering
\caption{Qualitative comparison of \algname{} ($K=4$) versus a naive $2\times2$ patch-based baseline on CIFAR-10 and CelebA. \algname{} maintains sharper details and avoids mosaic artifacts.}
\vspace{1mm}
\label{tab:grid-images}

\renewcommand{\arraystretch}{1.1}
\setlength{\tabcolsep}{2pt}

\resizebox{\linewidth}{!}{
\begin{tabular}{p{0.2475\linewidth} p{0.2475\linewidth} | p{0.2475\linewidth} p{0.2475\linewidth}}
\toprule
\multicolumn{2}{c}{\textbf{CIFAR-10}} &
\multicolumn{2}{c}{\textbf{CelebA}} \\
\midrule

\makebox[\linewidth][c]{\algname{} (K=4)} &
\makebox[\linewidth][c]{Patch-Based} &
\makebox[\linewidth][c]{\algname{} (K=4)} &
\makebox[\linewidth][c]{Patch-Based} \\

\midrule

\makebox[\linewidth][c]{\includegraphics[width=\linewidth]{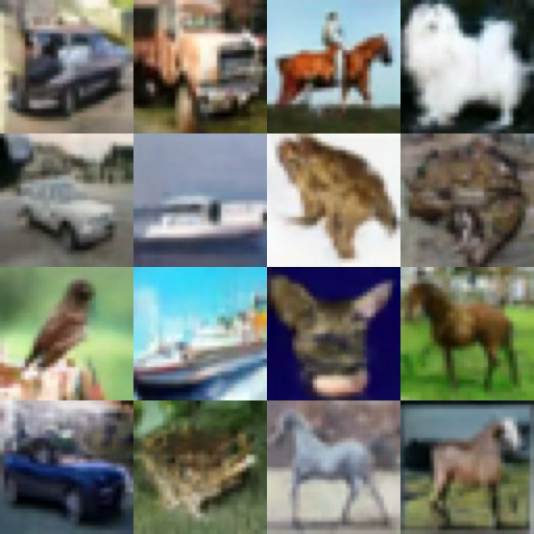}} &
\makebox[\linewidth][c]{\includegraphics[width=\linewidth]{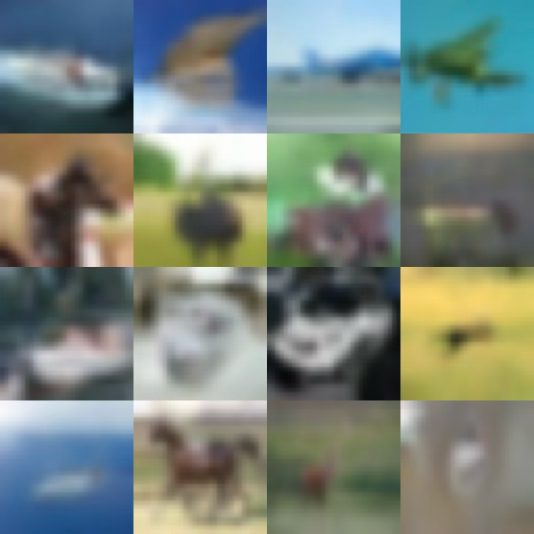}} &
\makebox[\linewidth][c]{\includegraphics[width=\linewidth]{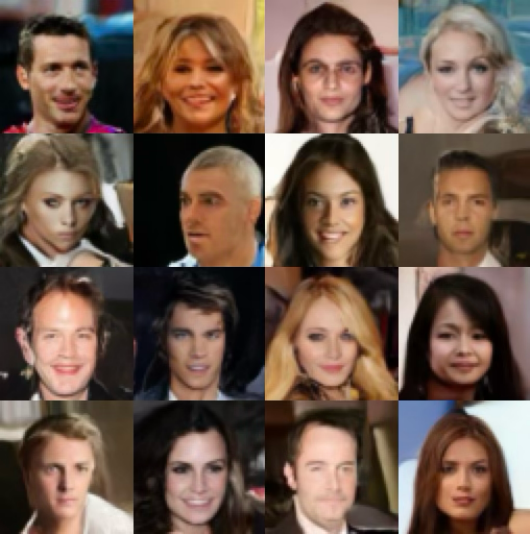}} &
\makebox[\linewidth][c]{\includegraphics[width=\linewidth]{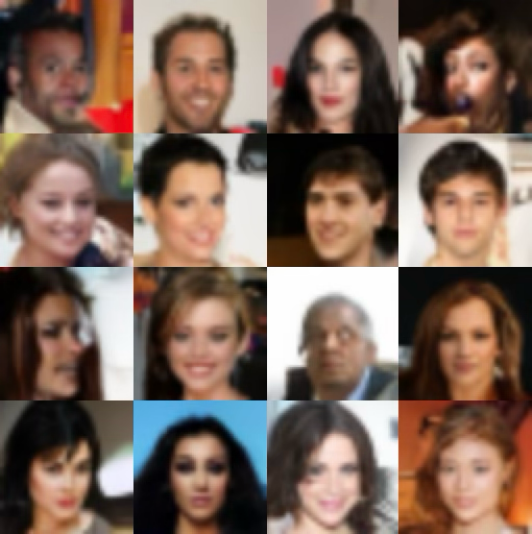}} \\

\bottomrule
\end{tabular}
}

\end{table}

\begin{table}[!htbp]
    \centering
    \caption{Quantitative performance of \algname{} (\(K=4\)) compared with a naive patch-based baseline on CIFAR-10 and CelebA. Metrics include FID, sFID, IS, Precision, and Recall. \algname{} significantly outperforms the naive approach.}
    \vspace{1mm}
    \label{tab:performance_appen}
    \begin{adjustbox}{max width=\linewidth}
      \begin{tabular}{l *{10}{c}}
          \toprule
          & \multicolumn{5}{c}{\textbf{CIFAR-10}} & \multicolumn{5}{c}{\textbf{CelebA}} \\
          \cmidrule(lr){2-6}\cmidrule(lr){7-11}
          Method
          & FID$\downarrow$
          & sFID$\downarrow$
          & IS$\uparrow$
          & Prec$\uparrow$
          & Rec$\uparrow$
          & FID$\downarrow$
          & sFID$\downarrow$
          & IS$\uparrow$
          & Prec$\uparrow$
          & Rec$\uparrow$ \\
          \midrule
          \algname{} (K=4)
            & 18.27 & 6.54 & 7.41 & 0.56 & 0.55
            & 9.50 & 10.66 & 2.58 & 0.70 & 0.38 \\
          Patch-Based
            & 91.71 & 52.16 & 4.55 & 0.35 & 0.07
            & 26.07 & 37.34 & 2.37 & 0.36 & 0.12 \\
          \bottomrule
      \end{tabular}
    \end{adjustbox}
\end{table}

\FloatBarrier
\paragraph{\textbf{Comparison with Capacity-Reduced Models.}}
Beyond comparisons to naive compositing, we also evaluate whether \algname{} simply mimics the effect of reducing model size to improve throughput. To this end, we construct two reduced-computation U-Net variants trained on CIFAR-10 at $K=2$.
The first is a \emph{channel-reduced} model, which decreases the number of feature maps per layer by modifying the channel multipliers. The second is a \emph{depth-reduced} model, which shortens the network depth by removing residual blocks. These variants allow us to compare shared computation with reductions in backbone capacity at comparable computational budgets.

Detailed configurations are listed in Table~\ref{tab:ddpm_cifar10_configs}, and quantitative results are reported in Table~\ref{tab:ddpm_cifar10_perf}. The reference in this comparison is \algname{} with the full backbone at $K=2$. Both reduced models require more MACs and achieve lower throughput than this reference. FID increases from 5.93 for \algname{} to 7.59 for the channel-reduced model and 6.55 for the depth-reduced model, with corresponding declines in IS and recall.

This comparison distinguishes two ways of reducing per-sample computation. Smaller backbones reduce the capacity used to process inputs, whereas \algname{} shares the computation of the full backbone across multiple samples. In the tested configurations, the latter achieves better quality at lower MAC counts and higher throughput. These results support retaining backbone capacity through fusion as an effective alternative to reducing channels or depth.

\begin{table}[!htbp]
    \centering
    \caption{Backbone configurations for the CIFAR-10 DDPM comparison between \algname{} and the channel-reduced and depth-reduced variants.}
    \vspace{1mm}
    \label{tab:ddpm_cifar10_configs}
    \begin{adjustbox}{max width=\linewidth}
    \begin{tabular}{lccc}
        \toprule
        & \multicolumn{3}{c}{\textbf{CIFAR-10 (DDPM)}} \\
        \cmidrule(lr){2-4}
        & Full Backbone & Channel-Reduced & Depth-Reduced \\
        \midrule
        Channels                 & 128                & 128                  & 128 \\
        Channel Multiplier       & 1,2,2,2            & 1,1,1,2              & 1,1,2 \\
        Number of Heads          & 1                  & 1                    & 1 \\
        Activation Function      & SiLU               & SiLU                 & SiLU \\
        Diffusion Steps          & 1000               & 1000                 & 1000 \\
        Noise Schedule           & linear             & linear               & linear \\
        Beta Start               & 0.0001             & 0.0001               & 0.0001 \\
        Beta End                 & 0.02               & 0.02                 & 0.02 \\
        Dropout                  & 0.1                & 0.1                  & 0.1 \\
        Learning Rate            & 2e-4               & 2e-4                 & 2e-4 \\
        Batch Size               & 128                & 128                  & 128 \\
        \bottomrule
    \end{tabular}
    \end{adjustbox}
\end{table}

\begin{table}[!htbp]
    \centering
    \caption{Performance of \algname{} ($K=2$) and the capacity-reduced variants on CIFAR-10 with DDPM. \algname{} achieves better FID, sFID, IS, precision, and recall at lower MAC counts and higher throughput in this comparison.}
    \vspace{1mm}
    \label{tab:ddpm_cifar10_perf}
    \begin{adjustbox}{max width=\linewidth}
    \begin{tabular}{l|ccccccc}
        \toprule
        \multicolumn{1}{c}{} & \multicolumn{7}{c}{\textbf{CIFAR-10 (DDPM)}} \\
        \cmidrule(lr){2-8}
        Method
          & {\makebox[1.6cm][c]{Throughput}}
          & {\makebox[1.6cm][c]{MAC}}
          & {\makebox[1.4cm][c]{FID$\downarrow$}}
          & {\makebox[1.6cm][c]{sFID$\downarrow$}}
          & {\makebox[1.4cm][c]{IS$\uparrow$}}
          & {\makebox[1.8cm][c]{Prec$\uparrow$}}
          & {\makebox[1.6cm][c]{Rec$\uparrow$}} \\
        \midrule
        \algname{} ($K=2$) & 3.17 & 3.07T & 5.93 & 4.58 & 8.67 & 0.63 & 0.58 \\
        \midrule
        Channel-Reduced  & 2.75 & 3.26T & 7.59 & 4.78 & 8.25 & 0.62 & 0.55 \\
        \midrule
        Depth-Reduced    & 2.54 & 3.72T & 6.55 & 4.65 & 8.47 & 0.63 & 0.57 \\
        \bottomrule
    \end{tabular}
    \end{adjustbox}
\end{table}

\clearpage
\section{Qualitative Results}
We organize the qualitative comparisons by evaluation setting. We first compare fusion factors across samplers and fusion designs for latent diffusion, then examine throughput scaling.

\paragraph{Samples across Samplers and Fusion Factors}
Tables~\ref{tab:image-grid-cifar} and~\ref{tab:image-grid-celeba} show random samples from DDPM, DDIM, and Flow Matching on CIFAR-10 and CelebA for $K=1,2,4$ and TFS. These comparisons complement the quantitative evaluation of sample quality across samplers and fusion factors.

\begin{table}[!htbp]
\centering
\caption{Randomly generated samples from DDPM, DDIM (T=50), and Flow Matching on \textbf{CIFAR-10} for \(K=1,2,4\) and TFS. \algname{} maintains coherence and diversity across different samplers.}
\label{tab:image-grid-cifar}

\renewcommand{\arraystretch}{1.1}
\setlength{\tabcolsep}{2pt}

\begin{tabular}{c c c c}
\toprule
K & DDPM & DDIM (T=50) & FM \\
\midrule

1 &
\includegraphics[width=0.29\linewidth]{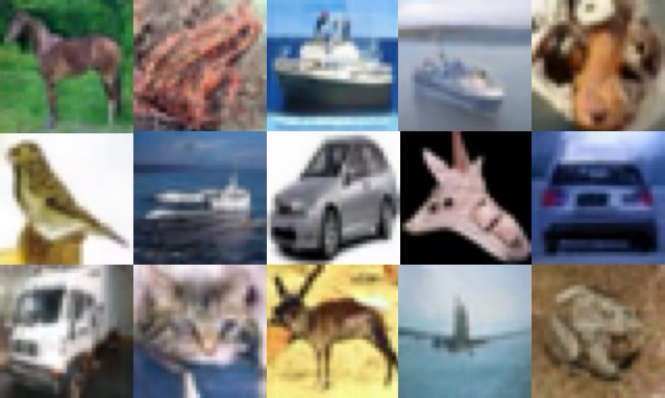} &
\includegraphics[width=0.29\linewidth]{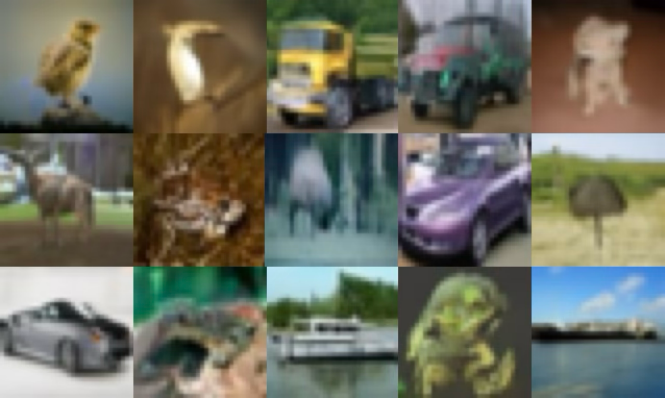} &
\includegraphics[width=0.29\linewidth]{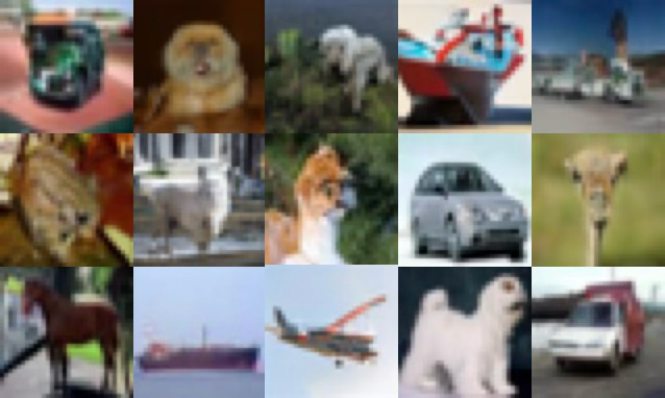} \\

2 &
\includegraphics[width=0.29\linewidth]{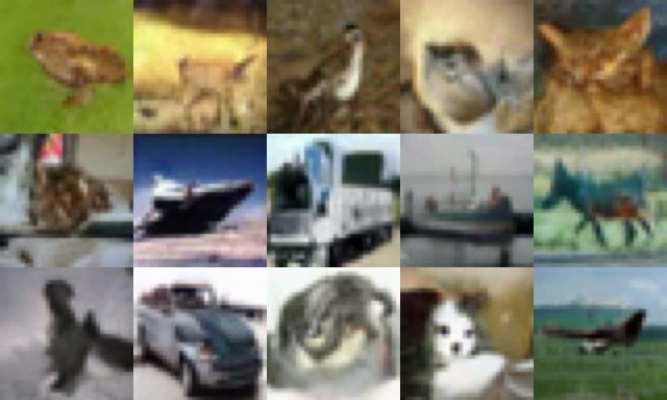} &
\includegraphics[width=0.29\linewidth]{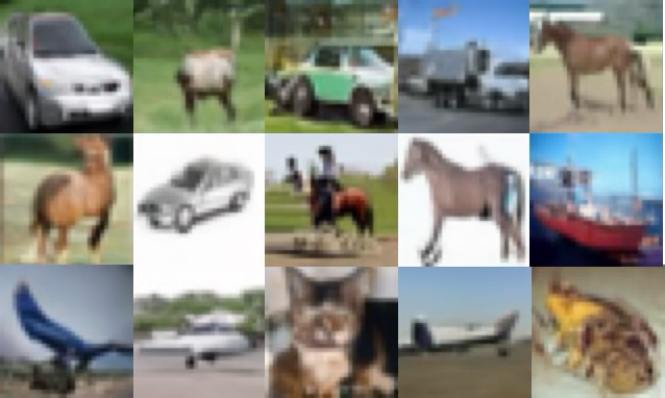} &
\includegraphics[width=0.29\linewidth]{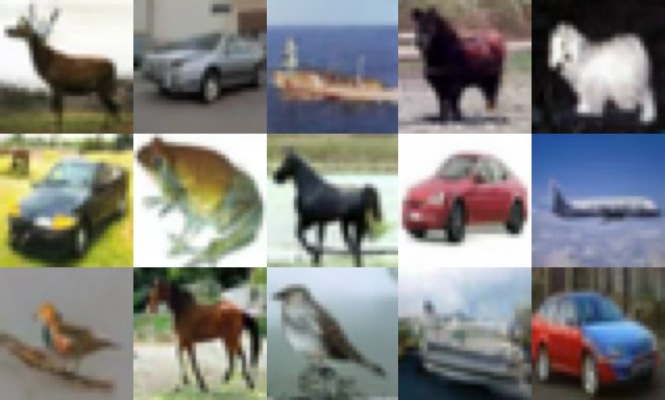} \\

4 &
\includegraphics[width=0.29\linewidth]{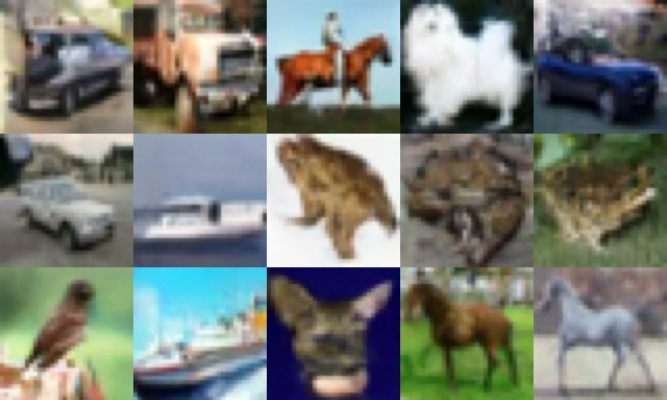} &
\includegraphics[width=0.29\linewidth]{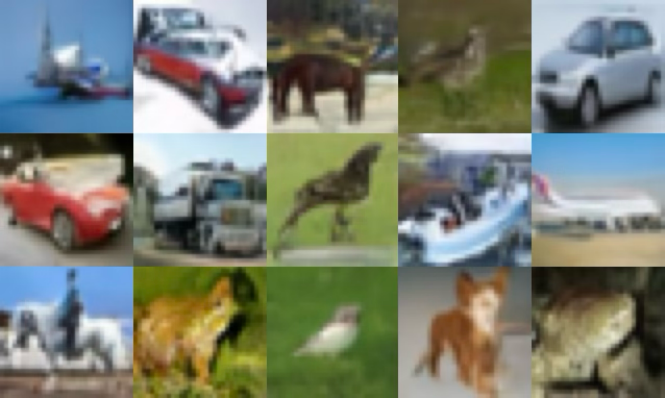} &
\includegraphics[width=0.29\linewidth]{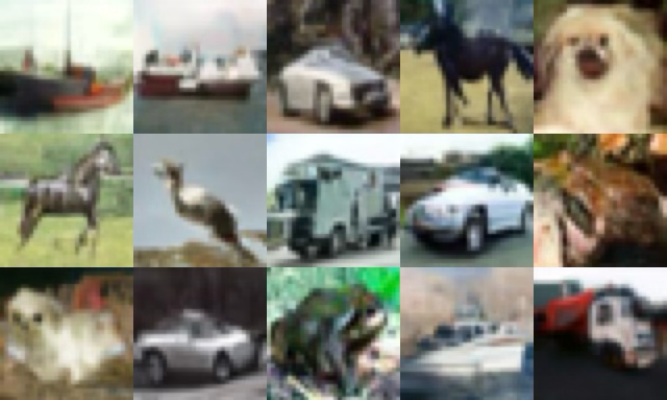} \\

TFS &
\includegraphics[width=0.29\linewidth]{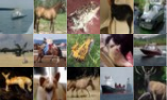} &
\includegraphics[width=0.29\linewidth]{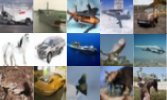} &
\includegraphics[width=0.29\linewidth]{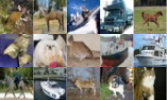} \\

\bottomrule
\end{tabular}

\end{table}

\begin{table}[!htbp]
\centering
\caption{Randomly generated samples from DDPM, DDIM (T=50), and Flow Matching on \textbf{CelebA} for \(K=1,2,4\) and TFS. Results confirm that \algname{} preserves sample fidelity and diversity across datasets.}
\label{tab:image-grid-celeba}

\renewcommand{\arraystretch}{1.1}
\setlength{\tabcolsep}{2pt}

\begin{tabular}{c c c c}
\toprule
K & DDPM & DDIM (T=50) & FM \\
\midrule

1 &
\includegraphics[width=0.29\linewidth]{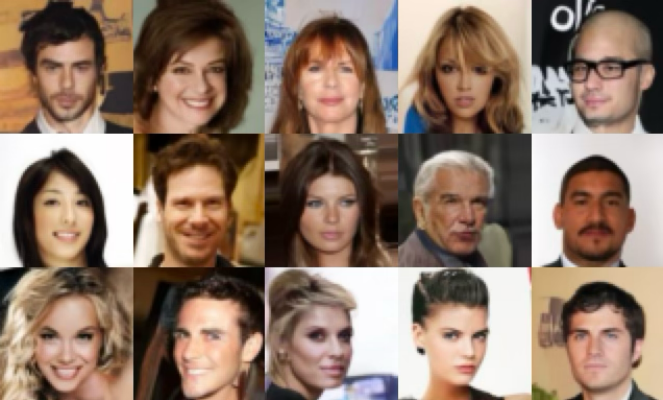} &
\includegraphics[width=0.29\linewidth]{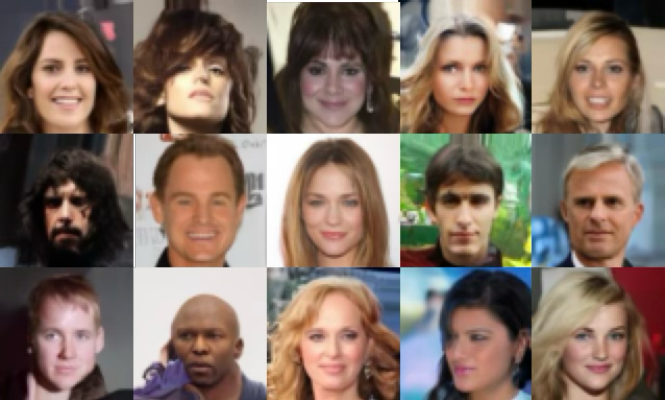} &
\includegraphics[width=0.29\linewidth]{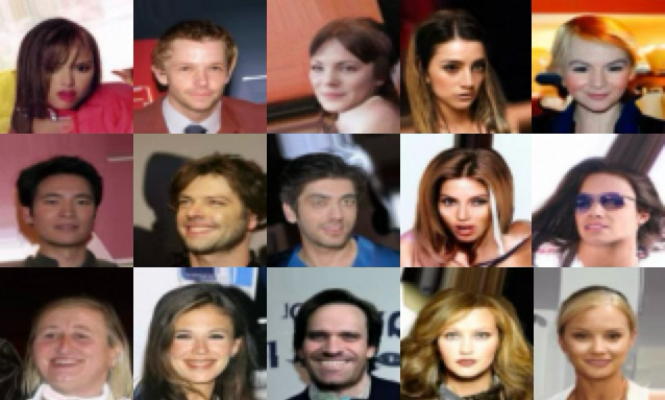} \\

2 &
\includegraphics[width=0.29\linewidth]{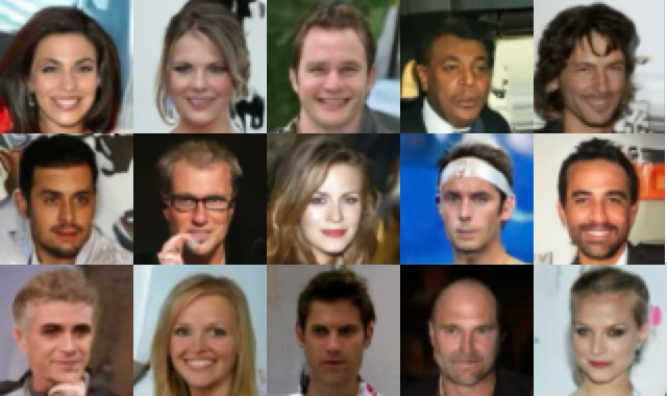} &
\includegraphics[width=0.29\linewidth]{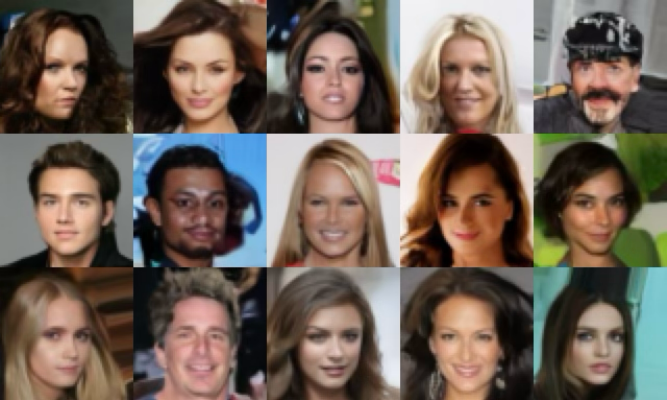} &
\includegraphics[width=0.29\linewidth]{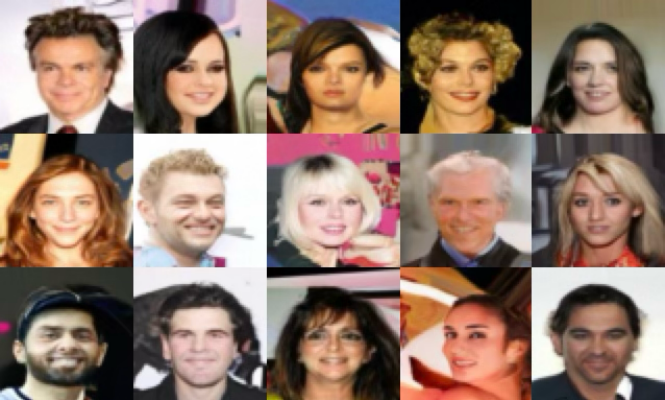} \\

4 &
\includegraphics[width=0.29\linewidth]{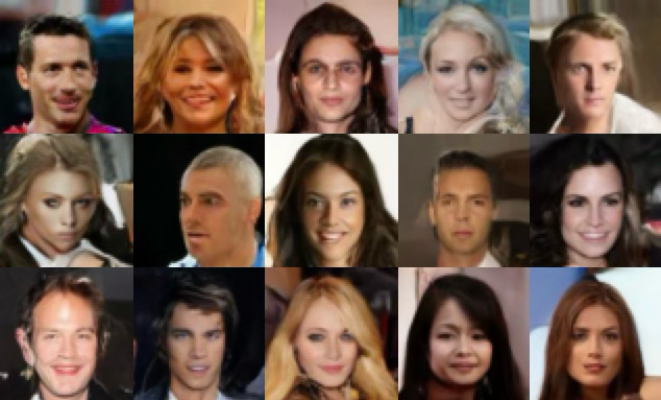} &
\includegraphics[width=0.29\linewidth]{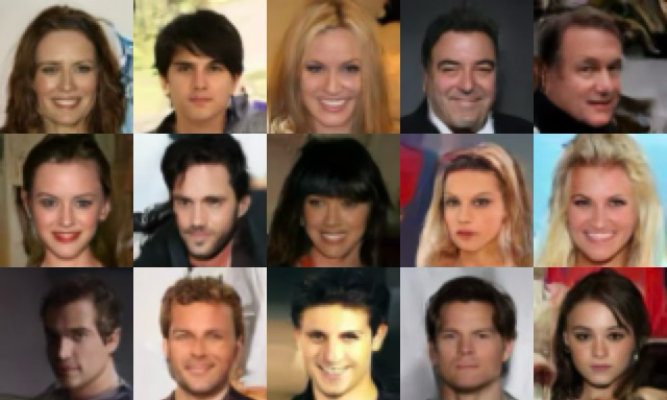} &
\includegraphics[width=0.29\linewidth]{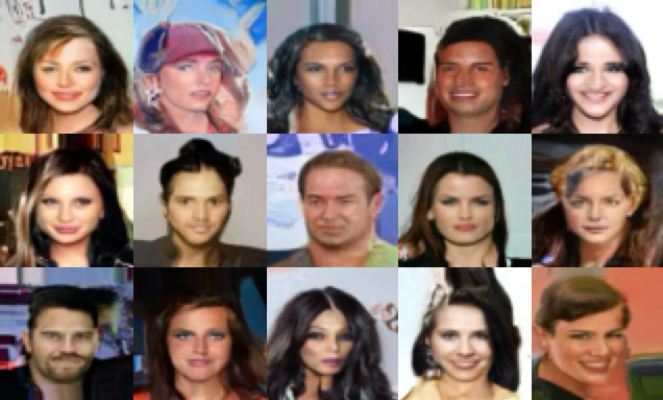} \\

TFS &
\includegraphics[width=0.29\linewidth]{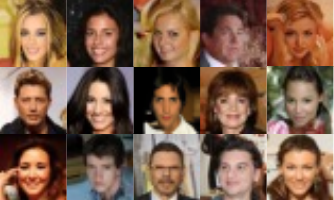} &
\includegraphics[width=0.29\linewidth]{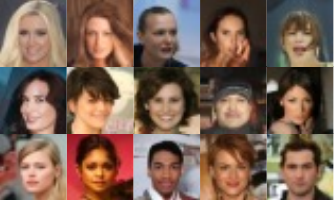} &
\includegraphics[width=0.29\linewidth]{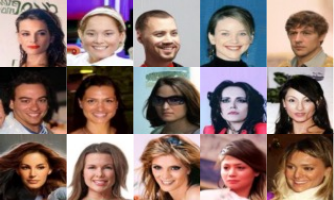} \\

\bottomrule
\end{tabular}
\end{table}

\FloatBarrier
\paragraph{Fusion Designs for Latent Diffusion}
Table~\ref{tab:ldm_comparison} compares the vanilla LDM with MF-AE and MF-UNet at $K=2$ and $K=4$, complementing the latent diffusion results above. For ImageNet, qualitative samples are omitted since the baseline DiT-S/2 configuration used in our experiments yields FID values above 50, rendering visual comparison uninformative.

\begin{table}[!htbp]
\centering
\caption{Qualitative comparison of vanilla LDM, MF-AE, and MF-UNet in latent diffusion models. Each fusion design is shown at $K=2$ and $K=4$.}
\label{tab:ldm_comparison}
\setlength{\tabcolsep}{2pt}
\begin{adjustbox}{max width=\linewidth}
\begin{tabular}{ccccc}
\toprule
LDM (Vanilla) & \multicolumn{2}{c}{MF-AE} & \multicolumn{2}{c}{MF-UNet} \\
\cmidrule(lr){2-3}\cmidrule(lr){4-5}
$K=1$ & $K=2$ & $K=4$ & $K=2$ & $K=4$ \\
\midrule
\includegraphics[width=0.18\linewidth]{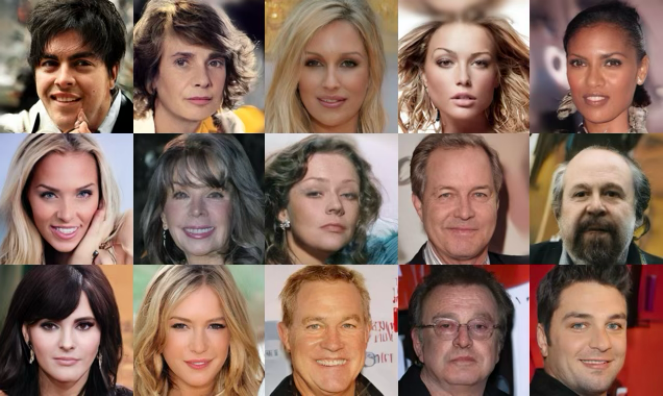} &
\includegraphics[width=0.18\linewidth]{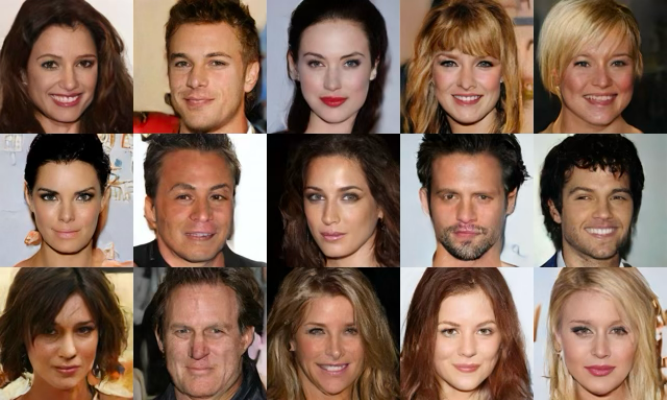} &
\includegraphics[width=0.18\linewidth]{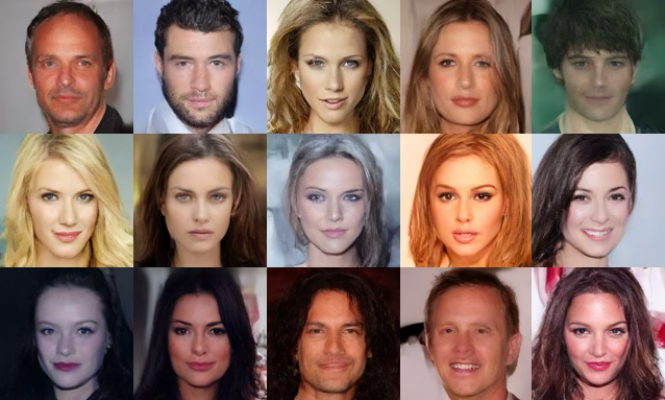} &
\includegraphics[width=0.18\linewidth]{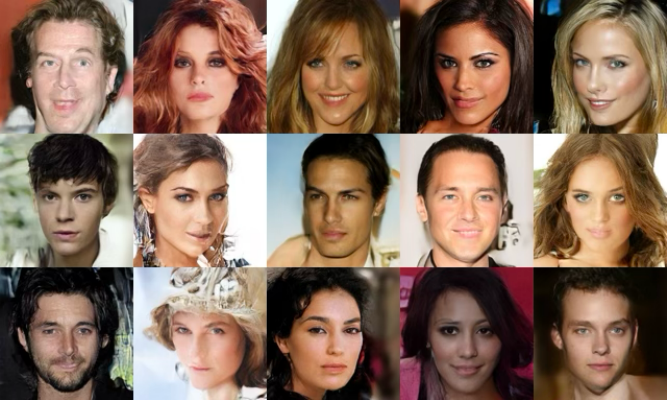} &
\includegraphics[width=0.18\linewidth]{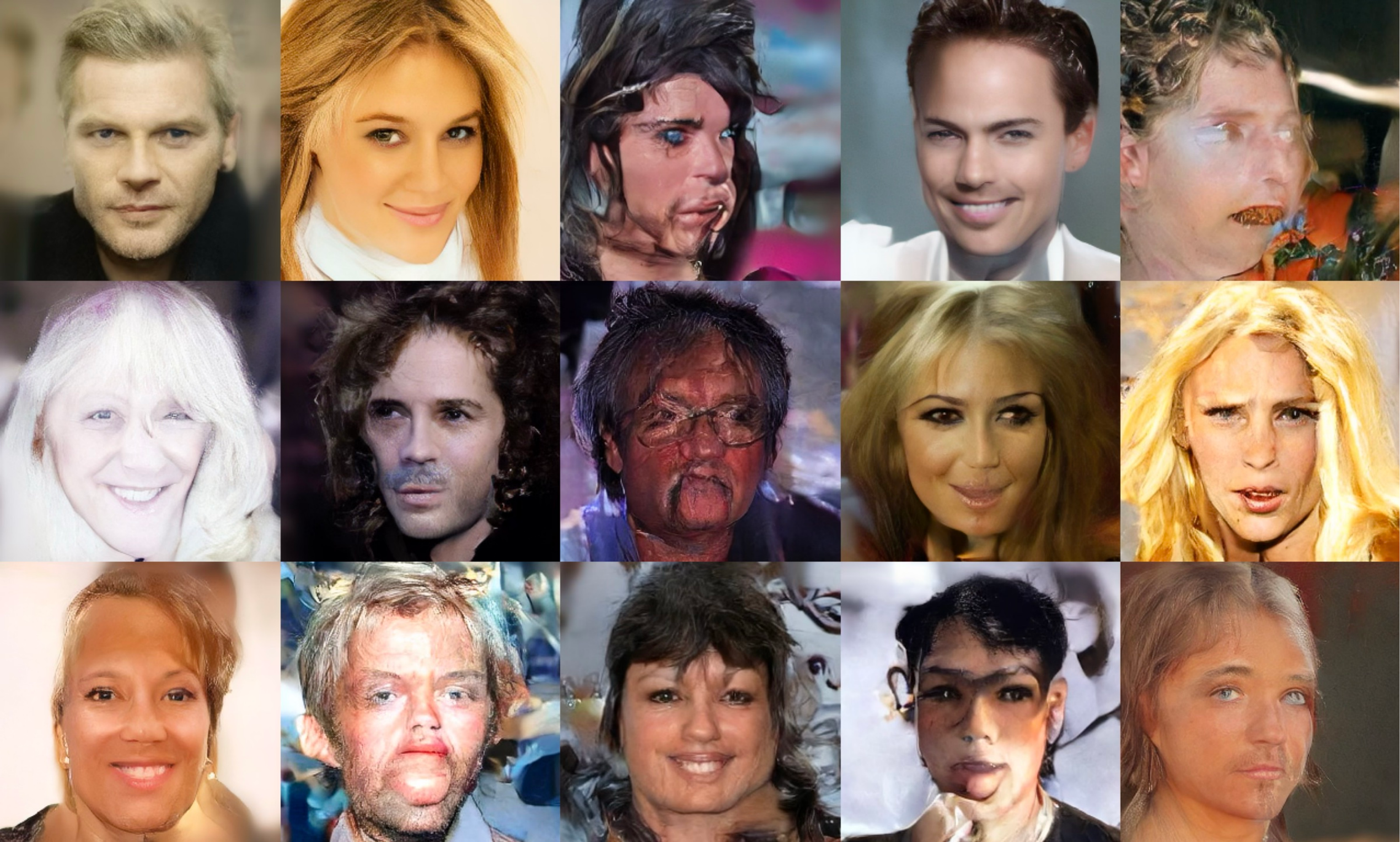} \\
\bottomrule
\end{tabular}
\end{adjustbox}
\end{table}

\clearpage
\paragraph{Quality across Throughput Levels}
Tables~\ref{tab:theoretical_samples_cifar} and~\ref{tab:theoretical_samples_celeba} compare the baseline, fixed fusion ($K=2$), and TFS at matched throughput levels on CIFAR-10 and CelebA. Columns follow the same order in both tables, from higher throughput to lower throughput.

\begin{table}[!htbp]
\centering
\caption{Qualitative scaling with throughput on \textbf{CIFAR-10}, showing random samples from \algname{} (K=2) and TFS using DDIM with timesteps 10, 20, 50, and 100. Throughput ranges from 66.80--565.03 img/s.}
\label{tab:theoretical_samples_cifar}

\renewcommand{\arraystretch}{1}
\setlength{\tabcolsep}{2pt}

\begin{adjustbox}{max width=0.9\linewidth}
\begin{tabular}{c c c c c}
\toprule
Throughput & 565.03\,img/s & 317.89\,img/s & 156.56\,img/s & 66.80\,img/s \\
\cmidrule(lr){1-5}

Baseline &
\includegraphics[width=0.21\linewidth,keepaspectratio]{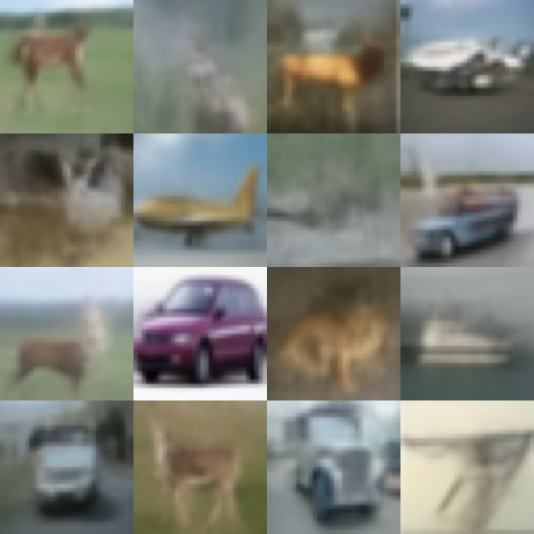} &
\includegraphics[width=0.21\linewidth,keepaspectratio]{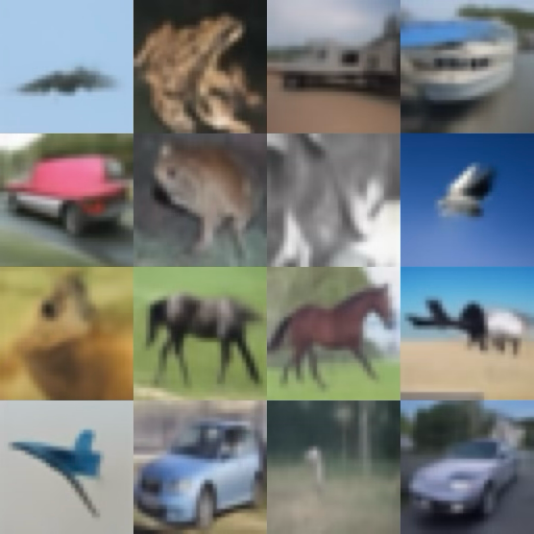} &
\includegraphics[width=0.21\linewidth,keepaspectratio]{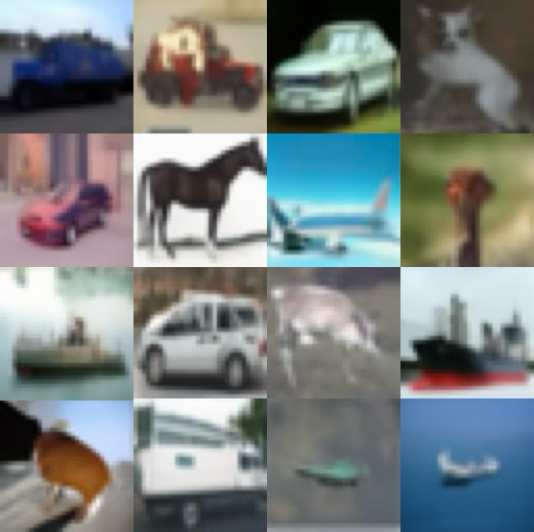} &
\includegraphics[width=0.21\linewidth,keepaspectratio]{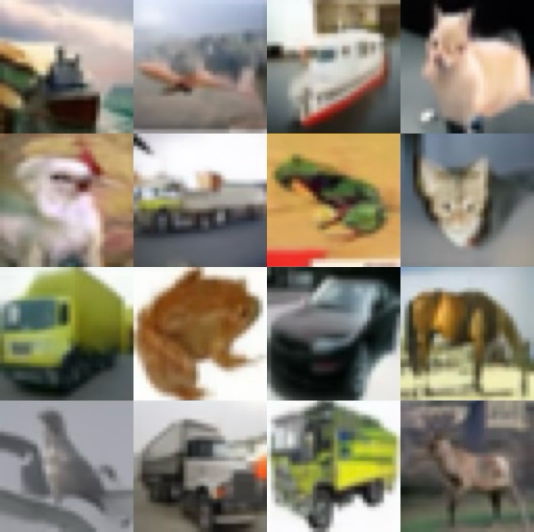} \\

\shortstack{\algname{}\\(K=2)} &
\includegraphics[width=0.21\linewidth,keepaspectratio]{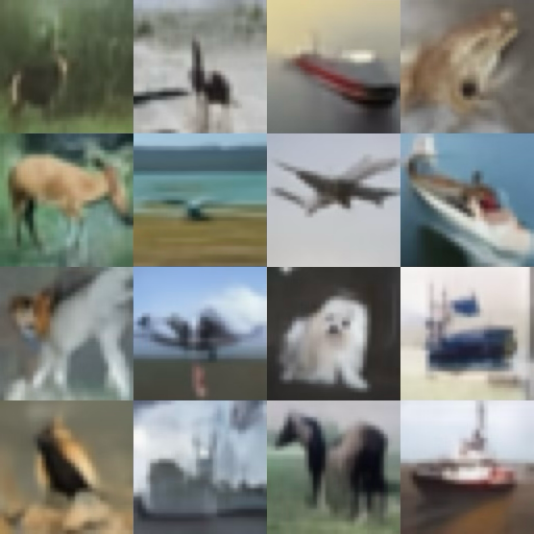} &
\includegraphics[width=0.21\linewidth,keepaspectratio]{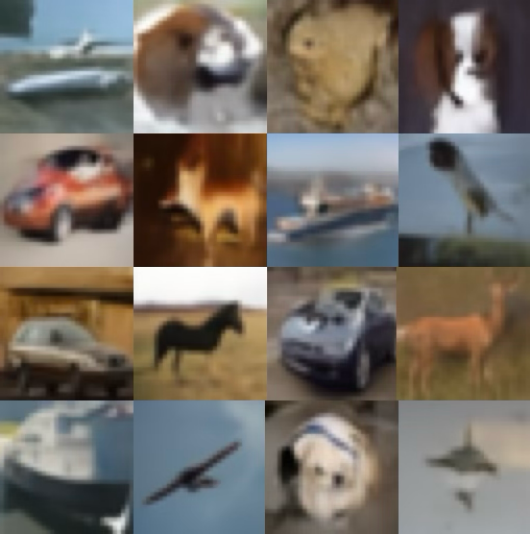} &
\includegraphics[width=0.21\linewidth,keepaspectratio]{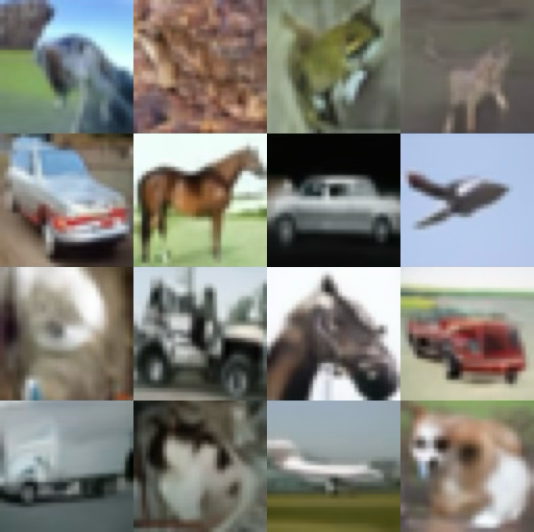} &
\includegraphics[width=0.21\linewidth,keepaspectratio]{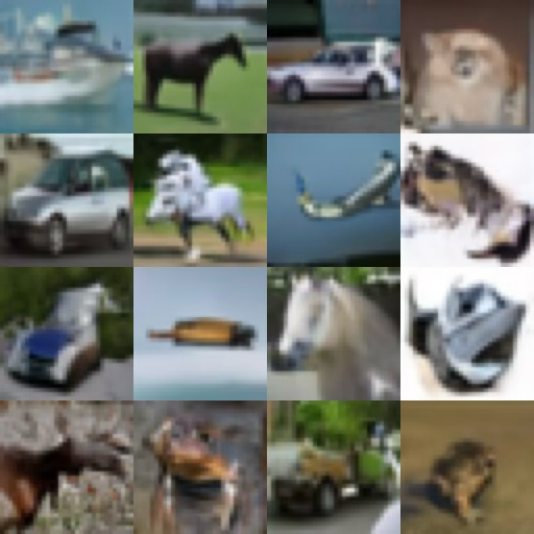} \\

TFS &
\includegraphics[width=0.21\linewidth,keepaspectratio]{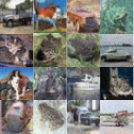} &
\includegraphics[width=0.21\linewidth,keepaspectratio]{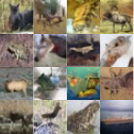} &
\includegraphics[width=0.21\linewidth,keepaspectratio]{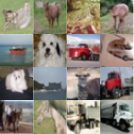} &
\includegraphics[width=0.21\linewidth,keepaspectratio]{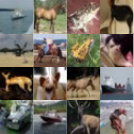} \\

\bottomrule
\end{tabular}
\end{adjustbox}
\end{table}

\begin{table}[!htbp]
\centering
\caption{Qualitative scaling with throughput on \textbf{CelebA}, showing random samples from \algname{} (K=2) and TFS using DDIM with timesteps 10, 20, 50, and 100. Throughput ranges from 26.83--228.77 img/s.}
\label{tab:theoretical_samples_celeba}

\renewcommand{\arraystretch}{1}
\setlength{\tabcolsep}{2pt}

\begin{adjustbox}{max width=0.9\linewidth}
\begin{tabular}{c c c c c}
\toprule
Throughput & 228.77\,img/s & 127.92\,img/s & 66.90\,img/s & 26.83\,img/s \\
\cmidrule(lr){1-5}

Baseline &
\includegraphics[width=0.21\linewidth,keepaspectratio]{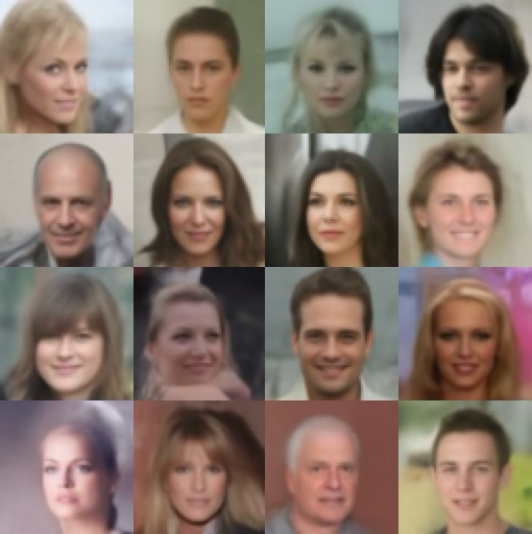} &
\includegraphics[width=0.21\linewidth,keepaspectratio]{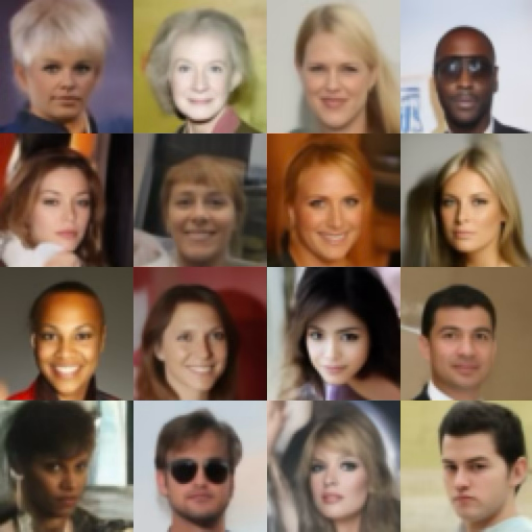} &
\includegraphics[width=0.21\linewidth,keepaspectratio]{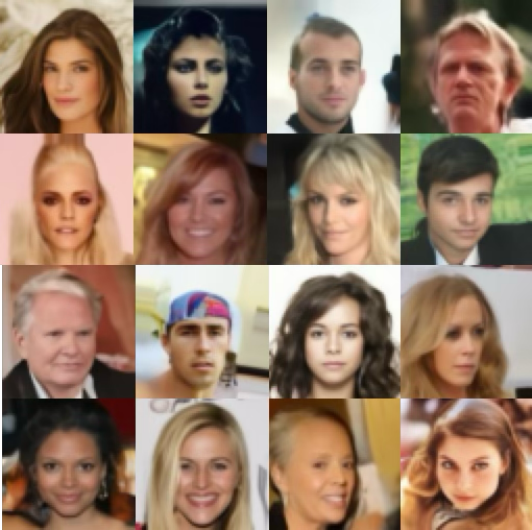} &
\includegraphics[width=0.21\linewidth,keepaspectratio]{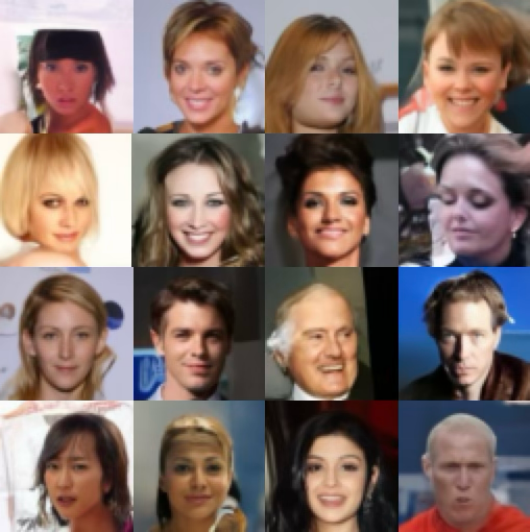} \\

\shortstack{\algname{}\\(K=2)} &
\includegraphics[width=0.21\linewidth,keepaspectratio]{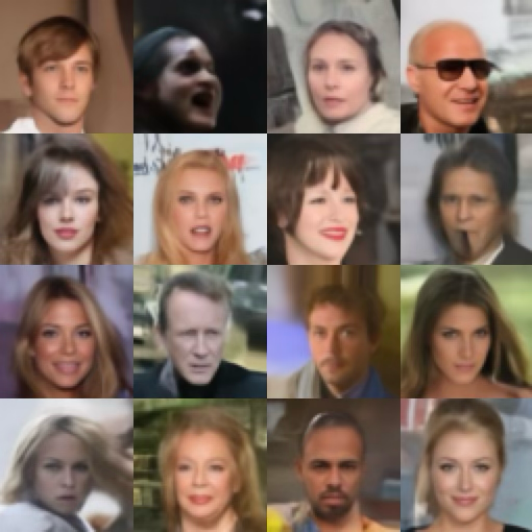} &
\includegraphics[width=0.21\linewidth,keepaspectratio]{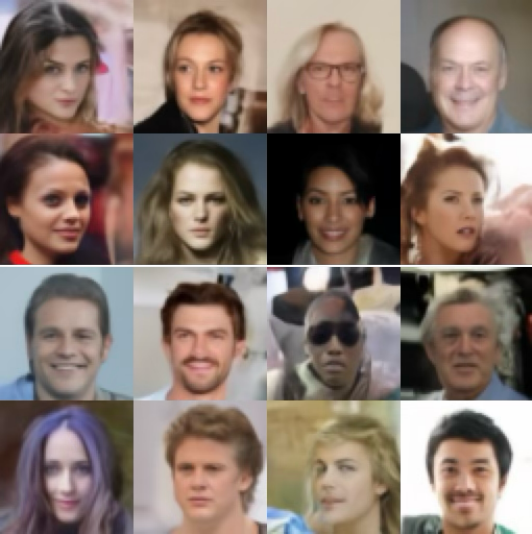} &
\includegraphics[width=0.21\linewidth,keepaspectratio]{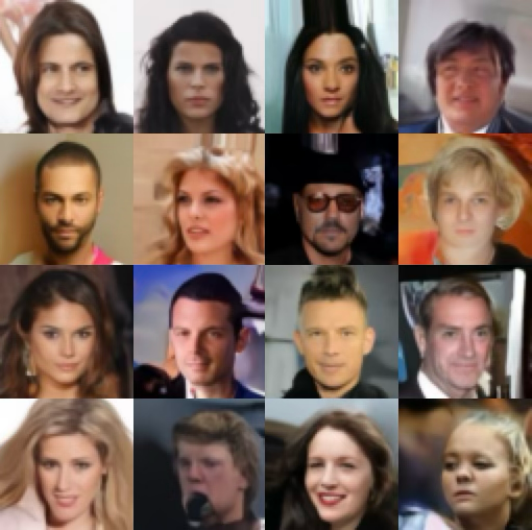} &
\includegraphics[width=0.21\linewidth,keepaspectratio]{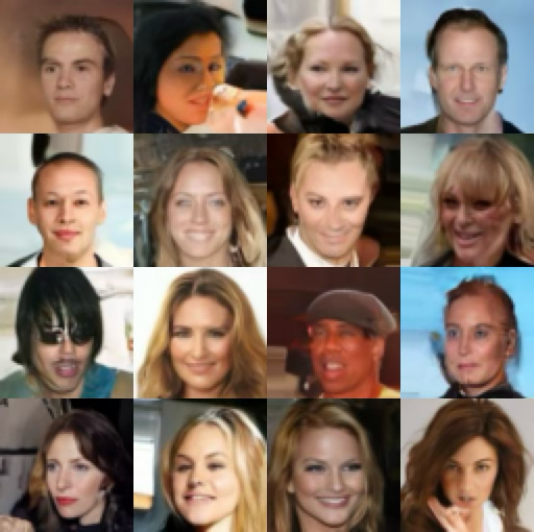} \\

TFS &
\includegraphics[width=0.21\linewidth,keepaspectratio]{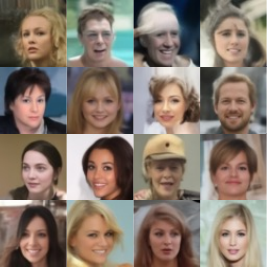} &
\includegraphics[width=0.21\linewidth,keepaspectratio]{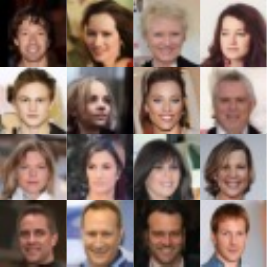} &
\includegraphics[width=0.21\linewidth,keepaspectratio]{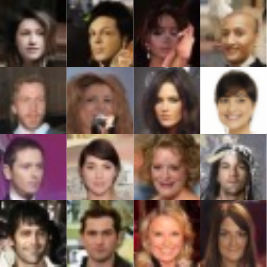} &
\includegraphics[width=0.21\linewidth,keepaspectratio]{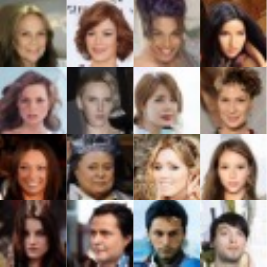} \\

\bottomrule
\end{tabular}
\end{adjustbox}
\end{table}

\clearpage

\newpage
\section*{NeurIPS Paper Checklist}

\begin{enumerate}

\item {\bf Claims}
    \item[] Question: Do the main claims made in the abstract and introduction accurately reflect the paper's contributions and scope?
    \item[] Answer: \answerYes{} %
    \item[] Justification: In both the abstract and the introduction, we underscored the limitations in prior works addressing the acceleration method of diffusion model inference. In response, we introduce an orthogonal approach to existing works.

\item {\bf Limitations}
    \item[] Question: Does the paper discuss the limitations of the work performed by the authors?
    \item[] Answer: \answerYes{} %
    \item[] Justification: The limitations of the work are discussed at the end of the conclusion (Section 6).

\item {\bf Theory assumptions and proofs}
    \item[] Question: For each theoretical result, does the paper provide the full set of assumptions and a complete (and correct) proof?
    \item[] Answer: \answerYes{} %
    \item[] Justification: The assumptions underlying the geometric analysis are stated in Section~\ref{subsec:geometric_analysis}, and the corresponding derivations and proofs are provided in Appendix~\ref{sec:target_details}. The experiments provide empirical evidence for the behavior discussed in the analysis.

    \item {\bf Experimental result reproducibility}
    \item[] Question: Does the paper fully disclose all the information needed to reproduce the main experimental results of the paper to the extent that it affects the main claims and/or conclusions of the paper (regardless of whether the code and data are provided or not)?
    \item[] Answer: \answerYes{} %
    \item[] Justification: The paper provides experimental settings and results across various frameworks. Table 1,2,3 include key informations necessary for reproducing experiments suach as model structure, number of parameters, and performance metrics.

\item {\bf Open access to data and code}
    \item[] Question: Does the paper provide open access to the data and code, with sufficient instructions to faithfully reproduce the main experimental results, as described in supplemental material?
    \item[] Answer: \answerYes{} %
    \item[] Justification: We provide code in supplementary.

\item {\bf Experimental setting/details}
    \item[] Question: Does the paper specify all the training and test details (e.g., data splits, hyperparameters, how they were chosen, type of optimizer) necessary to understand the results?
    \item[] Answer: \answerYes{} %
    \item[] Justification: The "Implementation Details" section explains how the models were implemented and metrics for performance evaluation are clearly defined both in Appendix.

\item {\bf Experiment statistical significance}
    \item[] Question: Does the paper report error bars suitably and correctly defined or other appropriate information about the statistical significance of the experiments?
    \item[] Answer: \answerNo{} %
    \item[] Justification: We did not perform repeated experimental runs because of computational constraints, so we do not report error bars, standard deviations, or statistical significance tests. Generative-quality evaluation uses 50k generated images per setting, but this sample count does not establish variability across independent runs.

\item {\bf Experiments compute resources}
    \item[] Question: For each experiment, does the paper provide sufficient information on the computer resources (type of compute workers, memory, time of execution) needed to reproduce the experiments?
    \item[] Answer: \answerYes{} %
    \item[] Justification: In appendix, we state that measurements are performed on a single NVIDIA A6000 GPU. Additionally, Table 1 provides detailed computational cost(MAC) and throughput for each model.
    
\item {\bf Code of ethics}
    \item[] Question: Does the research conducted in the paper conform, in every respect, with the NeurIPS Code of Ethics \url{https://neurips.cc/public/EthicsGuidelines}?
    \item[] Answer: \answerYes{} %
    \item[] Justification: The ManiFusion approach proposed in the paper is a conceptual methodology to improve inference efficiency of diffusion models and does not contain content that violates NeurIPS ethic guidelines. No issues related to privacy, social bias, or potential malicious use are existing.

\item {\bf Broader impacts}
    \item[] Question: Does the paper discuss both potential positive societal impacts and negative societal impacts of the work performed?
    \item[] Answer: \answerNo{} %
    \item[] Justification: The paper focuses on conceptual contributions and does not include a seperate discussion section on the positive/negative societal impacts of the proposed method.
    
\item {\bf Safeguards}
    \item[] Question: Does the paper describe safeguards that have been put in place for responsible release of data or models that have a high risk for misuse (e.g., pre-trained language models, image generators, or scraped datasets)?
    \item[] Answer: \answerNA{} %
    \item[] Justification: The paper proposes a methodology to improve the efficiency of existing diffusion model frameworks and does not release new models or datasets with misuse risks. Therefore, no special safeguards are necessary.

\item {\bf Licenses for existing assets}
    \item[] Question: Are the creators or original owners of assets (e.g., code, data, models), used in the paper, properly credited and are the license and terms of use explicitly mentioned and properly respected?
    \item[] Answer: \answerYes{} %
    \item[] Justification: All datasets are pubically available.

\item {\bf New assets}
    \item[] Question: Are new assets introduced in the paper well documented and is the documentation provided alongside the assets?
    \item[] Answer: \answerNA{} %
    \item[] Justification: The paper doesn't release new datasets, code or pre-trained models.

\item {\bf Crowdsourcing and research with human subjects}
    \item[] Question: For crowdsourcing experiments and research with human subjects, does the paper include the full text of instructions given to participants and screenshots, if applicable, as well as details about compensation (if any)? 
    \item[] Answer: \answerNA{} %
    \item[] Justification: This research is entirely algorithmic and computational in focus and does not involve human participants or crowdsourcing experiments.

\item {\bf Institutional review board (IRB) approvals or equivalent for research with human subjects}
    \item[] Question: Does the paper describe potential risks incurred by study participants, whether such risks were disclosed to the subjects, and whether Institutional Review Board (IRB) approvals (or an equivalent approval/review based on the requirements of your country or institution) were obtained?
    \item[] Answer: \answerNA{} %
    \item[] Justification: This research does not involve hunman subjects, so IRB approval or equivalent review is not required.

\item {\bf Declaration of LLM usage}
    \item[] Question: Does the paper describe the usage of LLMs if it is an important, original, or non-standard component of the core methods in this research? Note that if the LLM is used only for writing, editing, or formatting purposes and does \emph{not} impact the core methodology, scientific rigor, or originality of the research, declaration is not required.
    \item[] Answer: \answerNA{} %
    \item[] Justification: LLM is only used for writing and editing. There are no components in the development of the methodology in this research that were obtained from an LLM. 

\end{enumerate}

\end{document}